\documentclass{article}

\usepackage{arxiv_upload}

\usepackage[utf8]{inputenc} 
\usepackage[T1]{fontenc}

\usepackage{soul}

\usepackage{hyperref}       
\usepackage{url}            
\usepackage{booktabs}       
\usepackage{amsfonts}       
\usepackage{nicefrac}       
\usepackage{microtype}      
\usepackage{xcolor}         

\providecommand{\ywreply}[2]

\usepackage{times}
\usepackage{nicefrac}
\usepackage{mathtools}
\usepackage{amsthm}
\usepackage{amsmath}
\usepackage{amssymb}
\newcommand{\ie}{\emph{i.e.}}
\newcommand{\eg}{\emph{e.g.}}
\newcommand{\wrt}{\emph{w.r.t.}}

\newcommand{\expectation}{\mathbb{E}}

\newcommand{\argmax}{\text{argmax}}
\newtheorem{assumption}{Assumption}

\newtheorem{corollary}{Corollary}
\newtheorem{theorem}{Theorem}
\newtheorem{proposition}{Proposition}
\newtheorem{definition}{Definition}

\usepackage{graphicx}

\title{Tradeoffs Between Alignment and Helpfulness in Language Models with Steering Methods}
\author{Yotam Wolf, Noam Wies, Dorin Shteyman, Binyamin Rothberg, Yoav Levine, and Amnon Shashua
\\The Hebrew University
\\\texttt{\{yotamwolf,noam.wies,dorin.shteyman,binyamin.rothberg}\\\texttt{yoav.levine,shashua\}@cs.huji.ac.il}}

\date{October 2023}

\begin{document}

\maketitle

\begin{abstract}
Language model alignment has become an important component of AI safety, allowing safe interactions between humans and language models, by enhancing desired behaviors and inhibiting undesired ones.
It is often done by tuning the model or inserting preset aligning prompts.
Recently, \textit{steering methods}, such as representation engineering and feature steering and activation steering, methods which alters the model's behavior via changing its representations post-training, were shown to be effective in aligning LLMs. 
Steering methods yield gains in alignment oriented tasks such as resistance to adversarial attacks and reduction of social biases, but were also shown to  cause a decrease in 
the ability of the model to perform basic tasks. In this paper we study the tradeoff between the increase in alignment and decrease in helpfulness of the model. We propose a theoretical framework which provides bounds for these two quantities, and demonstrate their relevance empirically. First, we find that under the conditions of our framework, alignment can be guaranteed with steering methods, and at the same time that helpfulness is harmed in the process. Second, we show that helpfulness is harmed quadratically with the norm of the injected steering vectors, while the alignment increases linearly with it, indicating a regime in which it is efficient to use representation engineering. We validate our findings empirically, and chart the boundaries to the usefulness of these methods for alignment. Our code is available at \url{https://github.com/dorin133/REPE_alignment_helpfulness_tradeoff}.
\end{abstract}

\section{Introduction}

Advancements in large language model (LLM) development over the last few years have given LLMs a variety of abilities that allow them to serve as general purpose assistants in a wide range of tasks, such as broad-scoped question answering, 
writing assistance, teaching, and more
~\citep{Radford2019LanguageMA, devlin-etal-2019-bert, brown2020language,chatgpt,gpt4,bubeck2023sparks,nori2023capabilities,west2023advances,park2023generative}.
The vast use of LLMs for such purposes has raised concerns due to the harm they can cause their users, such as serving fake information~\citep{lin-etal-2022-truthfulqa,weidinger2022taxonomy}, behaving offensively, feeding social biases~\citep{hutchinson-etal-2020-social, venkit-etal-2022-study,weidinger2022taxonomy}, or encouraging problematic behaviors by users~\cite{nytimes_marry_reporter,suicide_convincing}. 
\textit{Alignment} is often the term given for the process of removing these undesired behaviors~\citep{yudkowsky2001creating,taylor2016alignment,amodei2016concrete,shalev2020ethics,hendrycks2021unsolved,pan2022the,ngo2022alignment}.

There are several different approaches to performing alignment in LLMs, such as including aligning prompts~\citep{askell2021general,rae2021scaling} which was shown to improve alignment and decrease toxicity in LLMs, and the procedure of reinforcement learning from human feedback (RLHF) which trains language models to be helpful and harmless~\citep{bai2022training}.
Though effective to an extent, these approaches are still dangerously frail, as
several works have shown that adversarial prompts can trigger negative behaviors in LLMs ~\cite{wallace-etal-2019-universal,yu-sagae-2021-automatically,xu-etal-2021-bot,subhash2023can,zou2023universal}. The work of \cite{wolf2023fundamental} provides a theoretical framework which shows that frozen LLMs can be misaligned with sufficiently long prompts.

Recently, new alignment methods were proposed, revolving around altering model weights at inference time, which control the model at the internal representations level by adding tailored vectors to the hidden layer's representations. The appeal of such methods is that enhancing concepts through finetuning is expensive and not always efficient for small changes, while inference time steering requires only inference compute and allows to specialize the model to the user's needs. Prominent methods include representation engineering \citep{zou2023representation} and activation steering \citep{turner2023activation}, in which directions in the model's latent space controlling certain behaviors are extracted by contrasting hidden representations in which opposing behaviors are exhibited,
as well as feature steering, by Anthropic \citep{templeton2024scaling}, in which steering vectors are obtained via the use of variational auto-encoders (VAEs), and demonstrate SOTA models such as Claude 3 Sonnet can be effectively steered by this method.
While the methods differ in their approach for obtaining the steering vectors, the underlying principle of injecting the vectors into the model is simlar.

Since then, there has been an increasing body of work using these methods. \cite{zou2023representation} demonstrated experimentally that the procedure can significantly
improve alignment, 
\eg, in resistance to adversarial attacks, with reduction from 50\% success of adversarial attacks to less than 15\%, and truthfulness enhancement, with a relative increase of over 50\%, though at the cost of somewhat reducing the helpfulness of the model. \citet{wang2024inferaligner} use extracted safety vectors for inference time alignment for harmlessness, reducing jailbreaking success rate from over 30\% with prompting and over 10\% in supervised fine tuning to below one percent. Similar methods have also been used by \cite{jorgensen2023improving,leong2023self,liu2023aligning,turner2023activation} to improve alignment and reduce toxicity. 
 \citet{wang2024detoxifying} uses a method of editing model parameters that maximize the difference between toxic and untoxic responses to detoxify it. \citet{wei2024assessing} find sparse regions in parameter space that affect alignment brittleness, to be removed for better alignment. \citet{marks2024sparse} interpret causal graphs in language models and edit them to improve behaviors. \cite{van2024extending} extend activation steering to multiple behaviors. To improve low rank finetuning, \cite{wu2024advancing} utilize a procedure of tuning representations directly to substantially reduce the trainable parameters of finetuning compared to LoRA. \cite{xu2024uncovering,li2024open} use concept activation vectors to jailbreak, they also observe that concepts that activate different behaviors are linearly separable. \cite{zhang2024truthx} remove hallucinations by editing truthfulness concepts. Additionally, the method scales to SOTA models, such as Claude 3 Sonnet \citep{templeton2024scaling}, using a similar method of sparse auto encoders, which extracts interpretable features from the model that can be used to manipulate the model through steering.
There are also known limitations to editing representations - \cite{yan2024potential} study limitations of model editing methods for social debiasing, and \cite{elazar2021amnesic} empirically demonstrate how projecting out supervised linear probe directions can reduce performance on selected tasks.

Understanding the tradeoff between model helpfulness and alignment is important for designing safe yet useful LLM systems. Previous empirical works have shown tradeoffs between quality and diversity and between helpfulness and safety in LLMs due to instruct finetuning \citep{florian2024exploring,bianchi2023safety,rottger2023xstest}, and reduction in performance due to watermarking \citep{ajith2023performance}.
In this work we aim to shed light on the benefits and limitations of steering for LLM alignment, \ie, how much does alignment improve with this method and what is the cost in terms of the model's abilities. We approach this question theoretically at first, and then provide empirical evidence for the validity of our theory.

\begin{figure}
    \centering
    
    \includegraphics[scale=0.45]{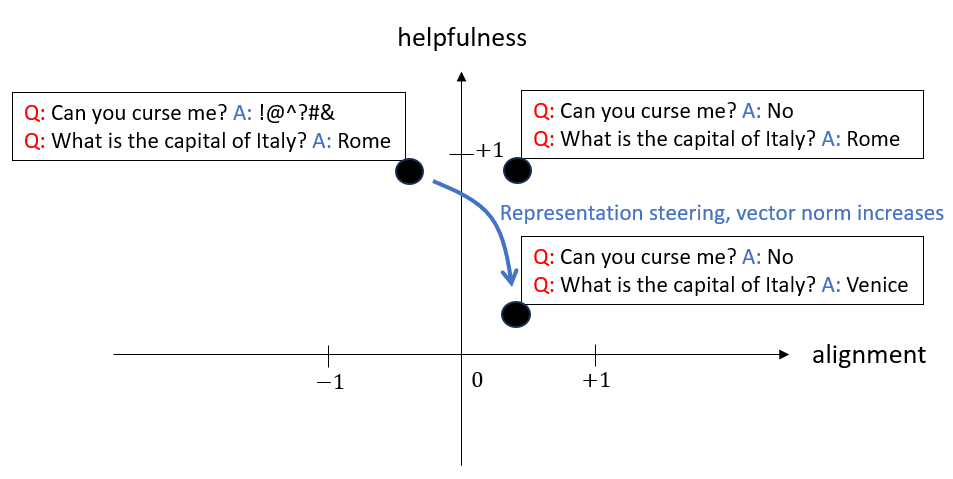}
    \caption{Effect of steering on helpfulness and alignment. Our main results show that alignment can improve at the cost of helpfulness. Moreover, we show that for small representation engineering norms the helpfulness decreases quadratically while the alignment increase is linear, so there is a regime in which representation engineering can be cost-effective. 
    }
    \label{fig:behavior_verticals}
\end{figure}

In sections \ref{sec:theoretical_framework} and \ref{sec:theoretical}, we set up our theoretical framework and present our theoretical results respectively. We find that steering increases alignment linearly with the steering vector norm (theorem \ref{theorem:1}), while the helpfulness of the model, defined as the probability of answering general queries correctly, decreases quadratically with the vector norm (theorem \ref{theorem:2}). Consequently, alignment can be guaranteed with large enough vector injections, 
though at the cost of reducing the model's helpfulness.
Conversely, when injecting vectors of small norms, the improvement of alignment is initially faster than the decrease in helpfulness, indicating a regime where steering is effective, allowing for inference time alignment while maintaining the model's helpful capabilities. See figure \ref{fig:behavior_verticals} for an illustration of this intuition. 

In section \ref{sec:exp} we explore the validity of our assumptions and results in an experimental setting with representation engineering: We calculate alignment, as defined by the theoretical framework, as a function of representation engineered vector norms corresponding to the desired behaviors and find that it increases as predicted by theorem \ref{theorem:1}. This is done by aligning with representation engineering an unaligned (pretrained) model with respect to desired behaviors (``harmless", ``not-racist"), and misaligning an aligned (RLHF) model to undesired behaviors (``harmful", ``racist"). Then, we calculate the helpfulness of the model, quantified by its question answering abilities over different knowledge domains and coding capabilities, with the same aligning vectors, and find that the decay with increased vector norm described in theorem \ref{theorem:2} is manifested. 
Together, the results correspond to the intuitive illustration in fig. \ref{fig:behavior_verticals}. Complementary experimental results by Anthropic showed similar empirical trends for alignment and helpfulness in the use of feature steering with vectors extracted from VAEs on Claude 3 Sonnet \citep{durmusevaluating}.

\section{Preliminaries}\label{sec:theoretical_framework}

We denote $P_\theta(\cdot|s)$ as the next token probability distribution of a model with parameters $\theta$, when conditioned  on the prompt $s$. The model is composed of $L$ layers, $r_\theta^{l}$ is the $l$'th hidden state representation of the model.
The next token prediction of a model is parametrized as:

    \begin{equation}
        P_\theta (t_{n+1}|t_1...t_n)=softmax(Ur^{(L)}_\theta(t_1...t_n))_{t_{n+1}}
    \end{equation}
    
Where $r^{(L)}_\theta(s)$ is the final hidden layer's representation of the prompt $s$ and $U$ is an unembedding matrix from the hidden state to a vocabulary of tokens, a standard parametrization for SOTA LLMs.
Denote a steered model by vectors, $R_e=(r_e^{(l=1)},...,r_e^{(l=L)})$, as $P_{\theta,r_e}$. Steering is performed at each layer by adding the corresponding vector to the hidden layer:

    \begin{equation}
         r_\theta^{(l)}\leftarrow r_\theta^{(l)} + r_e^{(l)} 
    \end{equation}

Additionally, we follow existing methods for steering and provide a uniform norm for all the injected vectors $|r_e^{(l)}| = |r_e|$, which are initially prepared with norm $1$, and when injected to the model, are multiplied by the coefficient $r_e$ which can be positive or negative, to tune the steering strength and direction. For layers that are not injected, $|r_e^{(l)}| = 0$.

To quantify alignment, we use the behavior expectation definition of alignment as in \cite{wolf2023fundamental}, based on the expected score of model responses to a behavior scoring function. The behavior scoring function can measure honesty, safety or any other concept for which responses can be scored as positively or negatively aligned with respect to. We will use a binary scoring function, with labels $\pm 1$ for aligned/misaligned answers. The results can be extended to more complex behavior scoring function over $[-1,+1]$, to yield qualitatively similar results, as discussed appendix \ref{sec:beyond_binary}:
\begin{definition}\label{def:behavior_expectation}
    Let $B:\Sigma^*\rightarrow\{-1,+1\}$ be a binary behavior scoring function, the behavior of a prompted model $P(\cdot|q)$ is defined as:
    \begin{equation}
        B[P_\theta(\cdot|q)] = \expectation_{a\sim P_\theta(\cdot|q)}[B(a)]=\sum_{a_+\in aligned}P_\theta(a_+|q)-\sum_{a_-\in misaligned}P_\theta(a_-|q)
    \end{equation}
\end{definition}

While $B$ is a binary function, the behavior expectation is in the range $[-1,+1]$, reflecting cases where a model has probability for both aligned and misaligned responses. In theorem \ref{theorem:1} we will prove that steering is an effective alignment method by lower bounding the behavior expectation. Notice that high probability of outputting a positive/negative response gives a positive/negative contribution to the behavior expectation, thus the sign and absolute value of behavior expectation measures the alignment of a model 
\wrt~the given behavior.

The model's helpfulness can be quantified as its ability to produce useful answers to user's queries (knowledge questions, code generation, summarization, etc.). In order to theoretically analyze helpfulness, we focus on queries where correctness can be defined, such as knowledge based question answering (see figure \ref{fig:behavior_verticals} for an example) and code generation. This can be measured as the likelihood of outputting a correct answer to a query:

\begin{equation}\label{def:helpfulness}
    helpfulness(model,q) =P_\theta(a_{correct}|q)
\end{equation}

Where $P_\theta(a_{correct}|q)$ is the model's probability of outputting the correct answer $a$ to the query $q$.
By this definition, the helpfulness is in the range $[0,1]$, in order to quantify the general capabilities of the model when steering vectors are injected into it.
For queries where correctness is not defined, the bounds we derive are expected to still be meaningful as they also describe the rate of the model's deviation from its original distribution due to steering.

The rational behind this quantification of alignment and helpfulness is to measure how aligning the model \wrt~a concept through steering affects its ability to perform other tasks. Ideally, a model that interacts with a user should be both aligned and helpful, meaning its response is appropriate \wrt~a desired behavior (quantified by a positive behavior expectation) and also useful (high probability of giving a correct answer to general purpose queries). 
In the next section, we will provide results on alignment and helpfulness under the use of steering, based on the model's next token prediction, which provides simple analytical forms for alignment and helpfulness. In appendix \ref{sec:multi_token}, we extend the results for multi-token answers, which yields qualitatively similar results, with somewhat more complex form.

\section{Main Results}\label{sec:theoretical}
We will show that steering improves alignment and harms helpfulness, yet a "moderate" use of steering can yield a model that is good for both. Theorem \ref{theorem:1} shows that behavior expectation is bounded from below by a hyperbolic tangent function, such that it approaches $+1$ for increasing size of injected vectors and increases linearly within a bounded range. This in principle allows to sample an aligned response for any adversarial attack (corollary \ref{cor:1}), demonstrating the power of representation engineering as an alignment technique. Theorem \ref{theorem:2} shows that the helpfulness is maximized in the vicinity of norm zero injected vectors (\ie, no representation engineering) and 
that as the norm is increased, helpfulness decays. The assumptions used to prove the theorems are presented formally in appendix \ref{sec:assumptions}.

\begin{figure}[h!]
    \centering
    \includegraphics[scale=0.45]{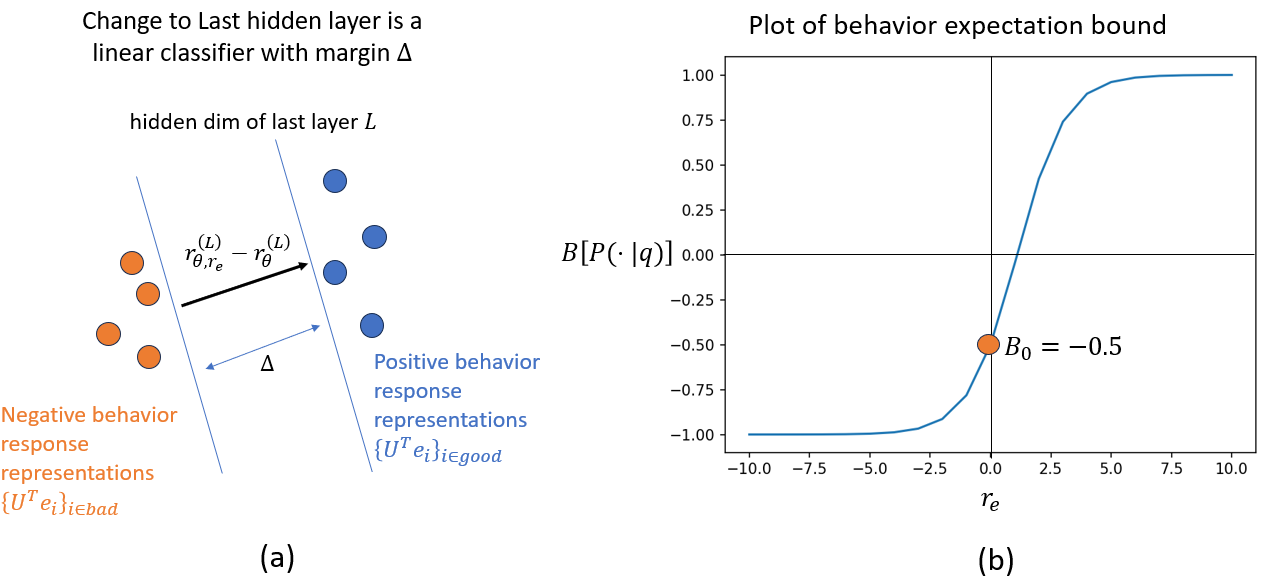}
    \caption{(a) The change to the last hidden layer due to vector injections from previous layers classifies positive and negative answer representations. (b) Plot of the upper bound on behavior expectation in theorem 
    \ref{theorem:1}.}
        \label{fig:behavior_expectation_bound}
\end{figure}

The following statement quantifies how alignment is improved by steering. It assumes the injected vectors in all layers accumulate to a change in the last hidden layer representation that classifies positive and negative behavior answers to the query, as depicted in figure \ref{fig:behavior_expectation_bound}a. This is assumed due to the popular choice in representation engineering to use steering vectors $\{r_e^{(l)}\}$, that are themselves classifiers for positive and negative representations on the intermediate layers, due to being learned from contrasting positive and negative behavior representations for different queries. For example, mean centering, $r_e^{(l)} = \expectation_{good,bad}[r_{good}^{(l)}-r_{bad}^{(l)}]$ (\cite{jorgensen2023improving}), or PCA, $r_e^{(l)}=\argmax_{v:||v||=1}[\expectation_{good,bad}|\langle v,r_{good}^{(l)}-r_{bad}^{(l)}\rangle|^2]$ (\cite{zou2023representation}),  such that they form linear classifiers for the intermediate layers due to the positive/negative inner product with positive/negative answer representations. Notably, in \cite{xu2024uncovering} it is shown empirically that such concept classes in latent space are linearly separable. We discuss this assumption further in \ref{sec:assumptions} and provide empirical evidence.
Furthermore, the classification condition can be softened to an imperfect classifier, as discussed in appendix \ref{sec:assumptions} and shown in appendix in \ref{sec:soft_margin}, to yield similar results.

\clearpage
\begin{theorem}\label{theorem:1}
Let $P_{\theta,r_e}(\cdot|q)$ be a model prompted with query $q$ and injected with representations of coefficient $r_e$. Let  $B:\Sigma^*\rightarrow \{-1,+1\}$ be a behavior scoring function.
The injections to all layers amounts to a change in the final hidden layer representation that is $q$ dependent, denoted by the vector $\delta r^{(L)}_e(q)$. Assume the representations of aligned and misaligned answers \wrt~$B$ are linearly separable, and $\delta r^{(L)}_e(q)$ linearly classifies them with margin $\Delta$. Then, the behavior expectation of the model conditioned on the query $q$ satisfies:
\begin{equation}
    B[P_{\theta,r_e}(\cdot|q)] \geq  tanh(\Delta \lambda \cdot r_e  + arctanh(B_0))
\end{equation}
Where $B_0 = B[P_\theta(\cdot|q)]$ is the behavior expectation without steering and $\lambda$ is a model dependent coefficient relating between $r_e$ and the corresponding final hidden state norm.
\end{theorem}
As can be seen in the mathematical expression and in figure \ref{fig:behavior_expectation_bound}b for $B_0=-0.5$, this lower bound is a shifted hyperbolic tangent function w.r.t $r_e$. At $r_e=0$ the bound gives $B_0$, which is the unaltered model's behavior. As $r_e$ is increased, the bound approaches $+1$, meaning the behavior asymptotically approaches $+1$. We see that for $B_0$ that is not too close to $-1$, the increase in behavior expectation is linear due to the hyperbolic tangent's nature, while if it is very close to $-1$, $r_e$ is to be increased before seeing the linear effect. Thus for behaviors on which the model is negative but has a small tendency for positive answers, the linear effect should be felt near $r_e=0$. In section \ref{sec:exp}, we present our numerical estimation $\Delta\lambda$ in the range $0.1-3$, both based on the linear classifier condition and direct alignment measurement. For proof see appendix section \ref{proof:theorem_1}.

This can be extended to multi-token answers, by enforcing the above result on each decoding step of the generated answer, as explained in appendix \ref{sec:multi_token}. 
The binary behavior score can also be extended beyond binary, as explained in appendix \ref{sec:beyond_binary}.
In contrast to \cite{wolf2023fundamental}, whose framework is centralized on using prompts to misalign frozen models, \ie~ whose weights and representations are not changed after training, here the model is not frozen due to steering, and accordingly a different result is obtained on guaranteeing an aligned response -- for any adversarial attack, using steering with large enough norms produces an aligned response if the learned steering vectors accumulate to a good classifier of positive and negative answer representations in the final layer. We formalize this in appendix \ref{section:corollary_1}.

Now, we shall bound from above the helpfulness of the model as a function of steering. We formally bound the probability of producing correct answers to queries where correctness is well defined. Yet, even when this is not the case, the bound can still be relevant, as it quantifies the model's deviation from its original distribution due to steering. Hence if the model was initially helpful on a task, a random deviation to its probability distribution is expected to decrease model performance proportionally to the size of the deviation.

Intuitively, editing the model's representation in a specific direction adds random noise to other latent concepts of the model, causing a degradation in its other capabilities. This is introduced in our framework through the resulting change to the final hidden layer $\delta r_e(q)=r_{\theta,r_e}^{(L)}-r_\theta^{(L)}$, we will assume its direction $\frac{\delta r_e(q)}{|\delta r_e(q)|}$ contains random projections \wrt~latent representations of correct and incorrect answers, which creates noise in the model's distribution. The noise is expected to be random on the highest probability tokens, when answering a query that is unrelated to the behavior being enhanced (intuitively depicted in figure \ref{fig:helpfulness_bound}a). We verify this empirically in appendix \ref{sec:exp_assumptions}. Thus, we assume random noise on the top $T$ tokens making up a large probability mass of the answer distribution, $1-\epsilon$, (\eg~$T\sim10$ typically makes $\epsilon\sim 0.1$), and do not make assumptions on the rest of the vocabulary. The following theorem formally states this.

\begin{theorem}\label{theorem:2}

    Let $P_{\theta,r_e}(\cdot|q)$ be a model prompted with query $q$ and injected with representations of coefficient $r_e$. If the resulting change to the directionality of the last hidden layer representation due to the injections in all layers, distributes randomly with variance $\sigma^2>0$ \wrt~the representations of correct and incorrect answers making up $1-\epsilon$ of the probability mass, the helpfulness of the model on the query is bounded with probability $1-\frac{2}{T}$ by:
    \begin{equation}\label{eq:helpfulness_bound}
        P_{\theta,r_e}(a_{correct}|q) \leq \frac{P_0}{P_0 + (1-P_0)\cdot \alpha(1-\epsilon)(1+\frac{\lambda^2\sigma^2\beta^2}{2} r_e^2)}
    \end{equation}
Where $P_0=P_{\theta,r_e=0}(\cdot|q)$ is the probability of answering correctly without steering, $T$ is the number of tokens making $1-\epsilon$ of the probability mass and $\alpha,\beta >0$ that depend on the query. $\lambda$ is a model dependent coefficient relating between $r_e$ and the corresponding final hidden state norm.
\end{theorem}

The proof is presented in appendix \ref{proof:theorem_2} and the assumption formally defined in appendix \ref{sec:assumptions}. The above bound is illustrated in figure \ref{fig:helpfulness_bound}b for different values of $\beta$. As can be seen, around $r_e=0$, the bound is parabolic, \ie~ the decrease is proportional to $-r_e^2$, obtained by expanding the bound near $r_e=0$. On the other hand, for large $r_e$, we see a decay to zero at a rate proportional to $r_e^{-2}$, obtained by expanding the bound for large $r_e$.
This result can be extended to multi-token answers, by enforcing the above result on each decoding step of the generated answer, as explained in appendix \ref{sec:multi_token}.

Importantly, this demonstrates that while large vector injections harm the model's overall performance, for small injections, the model's performance is relatively unharmed due to the slow (parabolic) decrease with norm around $r_e=0$. For the second statement to be feasible, the true helpfulness and the bound need to be close when no steering is performed. 
Indeed, the difference between the two at $r_e=0$ is bounded by $1-P_0$, such that for queries with high probability of being answered correctly without steering, \ie~$P_0\approx 1$, the true helpfulness and the bound will be close, guaranteeing the parabolic bound to be meaningful.

The parameter $\alpha\in[0,1]$ measures the tightness of the bound at $r_e=0$, since the true helpfulness at $r_e=0$ is $P_0$, while our helpfulness bound is $\frac{P_0}{P_0 + \alpha(1-P_0)}$. Thus $\alpha=1$ (and $\epsilon=0$) means the bound at $r_e=0$ coincides with the true helpfulness, while smaller $\alpha$ means the bound overshoots it. In our results, we obtain $\alpha\leq 0.5$. Figure \ref{fig:helpfulness_bound} depicts this overshooting for $\alpha=0.25$. Even so, as explained above, the tightness is at least $1-P_0$ regardless of $\alpha$, so it is always meaningful for queries the model is initially helpful on.

The product of parameters $\lambda\sigma\beta$ measures the rate/curvature of the quadratic decay, as they are the coefficient multiplying $r_e^2$. $\lambda$ is the same scaling parameter from theorem \ref{theorem:1}, $\sigma$ is the standard deviation of random noise added to the logits due to representation engineering (depicted in figure \ref{fig:helpfulness_bound}a and formally defined in \ref{sec:assumptions}). $\beta$ is the minimum between two weighted sums of positive variables with parameter $\sigma'=1$. In section \ref{sec:exp}, we present an empirical estimation for $\lambda\sigma\beta$ in the range $0.1-0.66$, based on the logit noise condition and direct helpfulness measurement. Hence the decay becomes strong at coefficients $r_e$ of size $1-10$.

\begin{figure}[h!]
    \centering
    \includegraphics[scale=0.45]{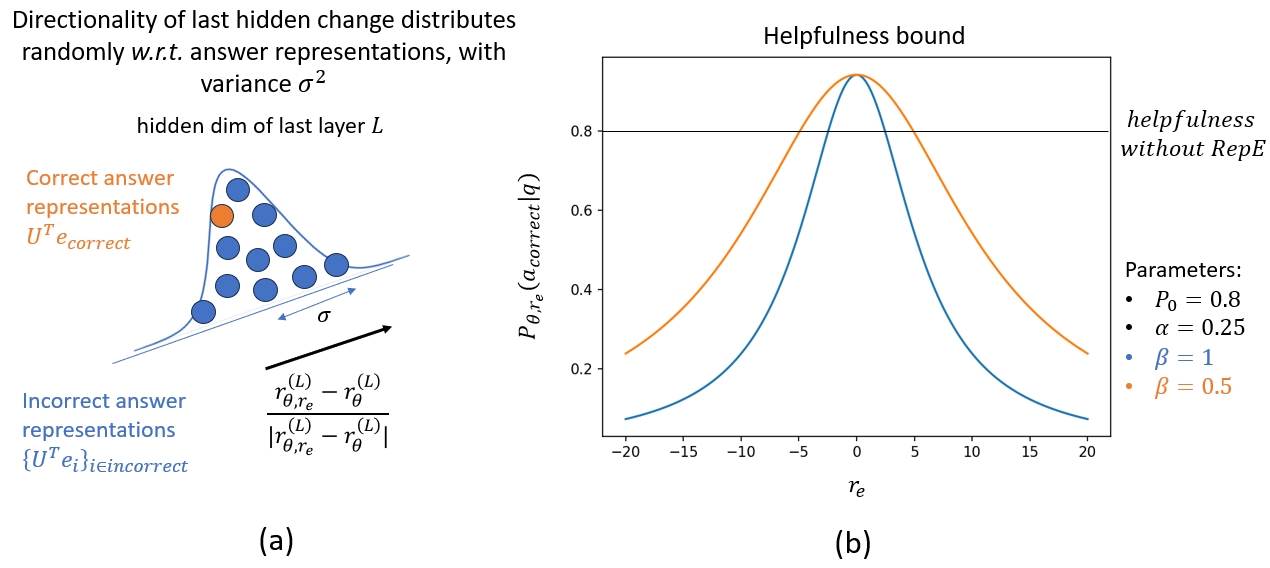}
    \caption{(a) Directionality of change to last hidden layer due to representation engineering distributes randomly with variance $\sigma^2$ \wrt~correct and incorrect answer representations. (b) Plot of helpfulness bound with given parameters of $P_0$, $\alpha$ and $\lambda\sigma\beta$.}
    \label{fig:helpfulness_bound}
\end{figure}

\paragraph{A tradeoff between alignment and usefulness:} The combination of the two results shows alignment improves linearly with the norm of the steering vectors while helpfulness is decreased quadratically. This means that when injecting vectors of small norms, the improvement of alignment is initially faster than the decrease in helpfulness, indicating a regime where steering is more effective. See figure \ref{fig:behavior_verticals} for an illustration of this intuition.

\section{Empirical results}\label{sec:exp} 

Here we will calculate alignment and helpfulness as defined above and observe how they are affected by increasing norms of steering vectors. Theorem  \ref{theorem:1} shows how alignment can increase/decrease due to steering, thus to demonstrate it, we increase the alignment of an unaligned pretrained model \wrt~ ``harmless" and ``not-racist" behaviors (specifically we use Llama 2 13B~\citep{touvron2023llama}), and conversely, misalign an aligned RLHF model \wrt~``harmful" and ``racist" behaviors (Llama 2 13B chat~\citep{touvron2023llama}).
Then, we calculate helpfulness as the probability of answering queries correctly when the model is injected with the same behavior altering vectors.
The experiments show an effect on alignment matching theorem \ref{theorem:1} and on helpfulenss matching theorem \ref{theorem:2}. Additional experimental details can be found in appendix \ref{sec:exp_details} as well as results for Llama 3.1 8B~\citep{dubey2024llama}. We note the goal of the experiments is to demonstrate the theoretical bounds showing an enhancement of alignment with a concept and a helpfulness decrease due to steering, and that a complementary experimental demonstration of these with more behaviors is shown on Claude 3 Sonnet with social biases when using feature steering \citep{durmusevaluating}.

We follow the work of \cite{zou2023representation} to extract the vectors used in representation engineering: Pairs of positive and negative statements \wrt~a behavior, are forward passed through the model, and the differences between representations of the pairs are used to find latent space directions that steer the model's responses from negative to positive behaviors or vice versa. For the ``harmful" behavior on the aligned model, we extracted harmful and unharmful instructions from AdvBench \citep{robey2021adversarial,robey2022probabilistically} and shareGPT respectively. For ``harmless" behavior on the unaligned model, the approach of contrasting positive and negative requests does not work, as the model agrees to answer both types of requests, so contrasting them does not steer the model towards not answering a request. Instead, inspired by the method of preference learning, we contrast aligned and misaligned responses to harmful instructions from AdvBench. For ``racism" on the aligned model, we used biased and unbiased statements from the StereoSet dataset \citep{nadeem2020stereoset}. For ``not-racist" on the unaligned model, we used the racist statements from above, followed by aligned and misaligned responses. The obtained vectors were used to calculate behavior expectation and helpfulness of the model as the norm of the vectors increased.

\paragraph{Alignment Measurement: }To calculate harmful behavior expectation, we sampled full responses to harmful instructions and used the behavior scoring function that assigns an answer $B(answer)=\pm 1$ if the model answers a harmful instruction or refuses to and calculated its expectation value, which is the difference between probabilities of fulfilling and not fulfilling the instruction.
To calculate the racism behavior expectation, sampled full responses to racist statements and used a behavior scoring function that assigns an answer $B(answer)=\pm 1$ to agreeing/disagreeing with a racist statement, and calculated the expectation value of this function \wrt~ the model distribution, which is the difference in probabilities of agreeing and disagreeing with a racist statement.

\begin{figure}[h!]
    \centering
    \includegraphics[scale=0.5]{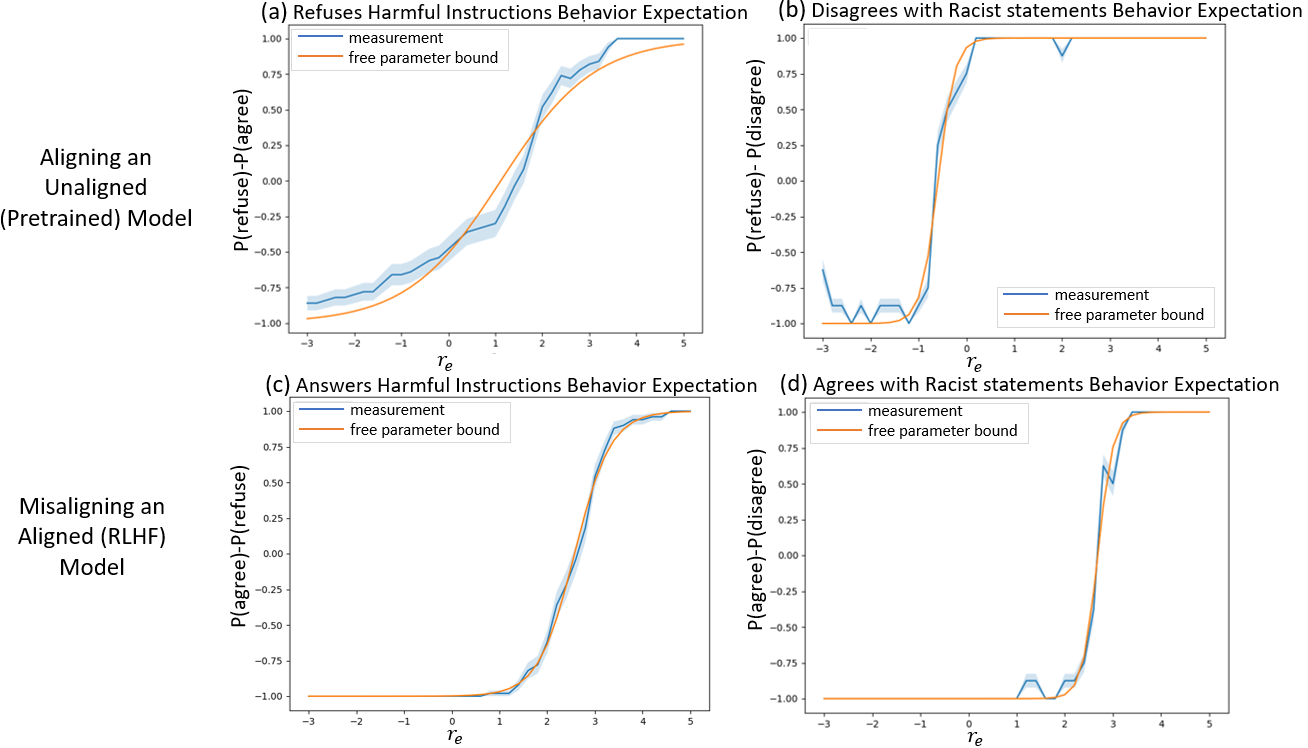}
    \caption{Plots of behavior expectation as a function of the coefficients of representation engineering vectors injected to the model. The blue line is the direct measurement, the orange line is a plot of the bound from theorem \ref{theorem:1}. (a) Harmless behavior expectation of Llama 2 13B as a function of coefficient of injected harmful PCA vectors. (b) Racism behavior expectation of Llama 2 13B as a function of coefficient of injected bias PCA vectors.(c) Harmful behavior expectation of Llama 2 13B as a function of coefficient of injected harmful PCA vectors. (d) Racism behavior expectation of Llama 2 13B chat as a function of coefficient of injected bias PCA vectors. }
\label{fig:behavior_expectation}
\end{figure}

Figure \ref{fig:behavior_expectation} shows behavior expectation as a function of corresponding PCA vector coefficients injected into the models. 
Overall we see that on both behaviors and both models, the behavior expectation changes like a hyperbolic tangent, as expected of theorem \ref{theorem:1}, which can be seen by the fitted curve of the data to a bound of the form of theorem \ref{theorem:1} when using $\Delta \lambda$ as a free parameter that fits the measurements. The value of $\Delta \lambda$ corresponding to the curve is $0.5-3$ while our empirically estimated value of $\Delta \lambda$ from the data based on the linear classification condition of the last hidden layer change is $0.1-0.4$ (for details and explanation for these differences see appendix \ref{sec:exp_assumptions}).
We note that for all behaviors, $r_e = 2.5$ suffices for a significant change in behavior expectation, taking it from negative to positive. It is left to observe the decrease in helpfulness and verify that it is not too big.

\paragraph{Helpfulness Measurement:}To calculate helpfulness, we tested the model on two tasks. The first is knowledge based question answering, for a clean test of the single token theoretical results (theorem \ref{theorem:2}). The second is code generation, to verify the single token results persist for tasks with multiple-token answers. Importantly, we injected the model with the same vectors used to alter the model's behavior in the alignment measurement.

\begin{figure}[h!]
    \centering
    \includegraphics[scale=0.32]{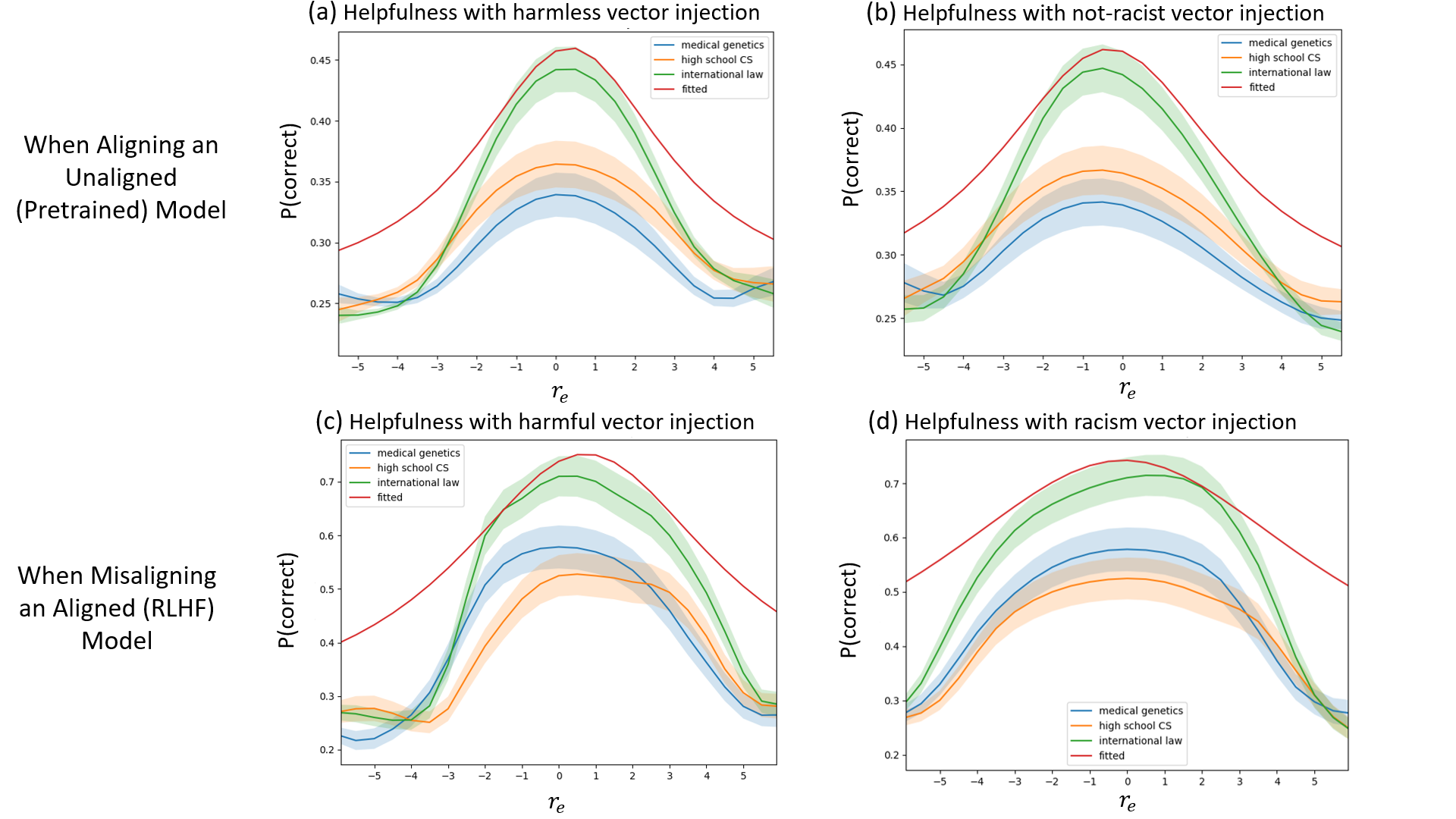}
    \caption{Helpfulness measurement: the probability assigned to the correct answer to questions from different MMLU tests (international law, medical genetics, high school computer science), as a function of representation engineering vector coefficients injected to the model. Here the probability of the correct answer was measured relative to the answers A, B, C, D. The red line plots the bound of theorem \ref{theorem:2} for free parameters on ``international law". (a) Helpfulness of Llama 2 13B with harmful PCA vectors. (b) Helpfulness of Llama 2 13B with bias PCA vectors. (c) Helpfulness of Llama 2 13B chat with harmful PCA vectors. (d) Helpfulness of Llama 2 13B chat with bias PCA vectors.}
    \label{fig:helpfulness_abcd}
\end{figure}

For the first task, we queried the model with multiple choice questions from the MMLU dataset \citep{hendrycks2020measuring} over a variety of domains (\eg~international law, medical genetics) and calculated the probability that the model assigns the correct answer. This was done both by calculating the probabilities of the multiple choice answers, A,B,C,D, and in appendix \ref{sec:exp_details} by sampling full responses to the questions and measuring the accuracy, yielding similar results. This was measured as a function of injected vector coefficients inserted to the model for the behaviors above.
Figure \ref {fig:helpfulness_abcd} shows the results for the different behaviors and models. 
We plot a bound of the form of theorem \ref{theorem:2} to demonstrate the predicted parabolic behavior.
We do so with free parameter $\lambda\sigma\beta$
from which we find $\lambda\sigma\beta$ in the range of $0.33$ to $0.66$ (see appendix \ref{helpfulness_details}). This is in accordance with our empirically estimated values of $0.1$ to $0.4$ for $\lambda\sigma\beta$ from direct measurement of the noise injected to the model due to representation engineering in appendix \ref{sec:exp_assumptions}.
Notably, for $r_e=2.5$, the decrease in helpfulness is still not too great, while as mentioned previously, alignment is significantly increased.

For the second task, we tested the model's coding skills with the humaneval dataset \citep{chen2021codex}. We present the results in appendix \ref{section:human_eval}. 
The model's performance is peaked around $r_e=0$, and it decays parabolically ar $r_e$ increases, as predicted in theorem \ref{theorem:2}.

\section{Discussion}
In this work, we study the benefits of steering methods for LLM alignment from a theoretical perspective.
We find that increasing the magnitude of the vectors injected to the model leads to improved alignment; we theoretically quantify this improvement as linear in the vectors' magnitude, and validate our result empirically. 
A practical outcome of our result is a guarantee of alignment when using the representation engineering method. 
Such theoretical guarantees cannot be made without altering the model at inference time -- \citet{wolf2023fundamental} show that prompt based alignment methods can always be undone. Our result thus crystallizes an inherent advantage of steering over competing alignment methods.

On the other hand, our framework indicates a degradation of the model's general capabilities when steering is applied. We theoretically quantify this degradation to be parabolic in the injected vectors' magnitude, which puts a bound on the strength with which steering should be performed to keep the model reliable for different uses. While our theoretical bound is an upper bound on the helpfulness, we observe this parabolic behavior empirically as well.

While steering is an emerging field, editing interpretable features of models on the representation level in order to control them scales to SOTA models such as Anthropic's Claude 3 Sonnet \citep{templeton2024scaling,durmusevaluating}. In principle, our framework may be generalized for theoretically analyzing the effects of normal finetuneing on alignment and helpfulness, as it too amounts to a change in the LLM representations to maximize the likelihood of desired outputs. In particular, each step in preference learning is equivalent to steering with coefficient that equals to the learning rate
(see appendix \ref{preference}), and indeed similar tradeoffs have been observed for finetuning \citep{tan2024truly}.
However, we leave this for future work, as finetuning creates small changes to the model's representation at each training step on several behaviors, that sums to a large overall change, while steering takes a large step in one direction.
As a result, the change to the representations in a steering process on one behavior creates random noise on the others (assumption \ref{assumption:normal}), unlike a finetuning process where this does not necessarily happen. Hence in regards of maintaining helpfulness, finetuning has an advantage, however, steering does enjoy the benefit of an online controllable step size in the desired behavior for effective manipulation at inference time.

Overall, we hope that our theoretical work will shed light on the mechanism of steering, which constitutes a new interesting direction for language model alignment.

\section*{Acknowledgments}
This research was supported by the ERC (European Research Council) and the ISF (Israel Science Foundation).

\clearpage
\bibliography{main}

\begin{thebibliography}{61}
\providecommand{\natexlab}[1]{#1}
\providecommand{\url}[1]{\texttt{#1}}
\expandafter\ifx\csname urlstyle\endcsname\relax
  \providecommand{\doi}[1]{doi: #1}\else
  \providecommand{\doi}{doi: \begingroup \urlstyle{rm}\Url}\fi

\bibitem[Ajith et~al.(2023)Ajith, Singh, and Pruthi]{ajith2023performance}
Anirudh Ajith, Sameer Singh, and Danish Pruthi.
\newblock Performance trade-offs of watermarking large language models.
\newblock \emph{arXiv preprint arXiv:2311.09816}, 2023.

\bibitem[Amodei et~al.(2016)Amodei, Olah, Steinhardt, Christiano, Schulman, and Man{\'e}]{amodei2016concrete}
Dario Amodei, Chris Olah, Jacob Steinhardt, Paul Christiano, John Schulman, and Dan Man{\'e}.
\newblock Concrete problems in ai safety.
\newblock \emph{arXiv preprint arXiv:1606.06565}, 2016.

\bibitem[Askell et~al.(2021)Askell, Bai, Chen, Drain, Ganguli, Henighan, Jones, Joseph, Mann, DasSarma, et~al.]{askell2021general}
Amanda Askell, Yuntao Bai, Anna Chen, Dawn Drain, Deep Ganguli, Tom Henighan, Andy Jones, Nicholas Joseph, Ben Mann, Nova DasSarma, et~al.
\newblock A general language assistant as a laboratory for alignment.
\newblock \emph{arXiv preprint arXiv:2112.00861}, 2021.

\bibitem[Atillah(2023)]{suicide_convincing}
Imane~El Atillah.
\newblock Man ends his life after an ai chatbot 'encouraged' him to sacrifice himself to stop climate change.
\newblock \emph{Euronews}, 2023.

\bibitem[Bai et~al.(2022)Bai, Jones, Ndousse, Askell, Chen, DasSarma, Drain, Fort, Ganguli, Henighan, et~al.]{bai2022training}
Yuntao Bai, Andy Jones, Kamal Ndousse, Amanda Askell, Anna Chen, Nova DasSarma, Dawn Drain, Stanislav Fort, Deep Ganguli, Tom Henighan, et~al.
\newblock Training a helpful and harmless assistant with reinforcement learning from human feedback.
\newblock \emph{arXiv preprint arXiv:2204.05862}, 2022.

\bibitem[Bianchi et~al.(2023)Bianchi, Suzgun, Attanasio, R{\"o}ttger, Jurafsky, Hashimoto, and Zou]{bianchi2023safety}
Federico Bianchi, Mirac Suzgun, Giuseppe Attanasio, Paul R{\"o}ttger, Dan Jurafsky, Tatsunori Hashimoto, and James Zou.
\newblock Safety-tuned llamas: Lessons from improving the safety of large language models that follow instructions.
\newblock \emph{arXiv preprint arXiv:2309.07875}, 2023.

\bibitem[Brown et~al.(2020)Brown, Mann, Ryder, Subbiah, Kaplan, Dhariwal, Neelakantan, Shyam, Sastry, Askell, et~al.]{brown2020language}
Tom Brown, Benjamin Mann, Nick Ryder, Melanie Subbiah, Jared~D Kaplan, Prafulla Dhariwal, Arvind Neelakantan, Pranav Shyam, Girish Sastry, Amanda Askell, et~al.
\newblock Language models are few-shot learners.
\newblock \emph{Advances in neural information processing systems}, 33:\penalty0 1877--1901, 2020.

\bibitem[Bubeck et~al.(2023)Bubeck, Chandrasekaran, Eldan, Gehrke, Horvitz, Kamar, Lee, Lee, Li, Lundberg, et~al.]{bubeck2023sparks}
S{\'e}bastien Bubeck, Varun Chandrasekaran, Ronen Eldan, Johannes Gehrke, Eric Horvitz, Ece Kamar, Peter Lee, Yin~Tat Lee, Yuanzhi Li, Scott Lundberg, et~al.
\newblock Sparks of artificial general intelligence: Early experiments with gpt-4.
\newblock \emph{arXiv preprint arXiv:2303.12712}, 2023.

\bibitem[Chen et~al.(2021)Chen, Tworek, Jun, Yuan, de~Oliveira~Pinto, Kaplan, Edwards, Burda, Joseph, Brockman, Ray, Puri, Krueger, Petrov, Khlaaf, Sastry, Mishkin, Chan, Gray, Ryder, Pavlov, Power, Kaiser, Bavarian, Winter, Tillet, Such, Cummings, Plappert, Chantzis, Barnes, Herbert-Voss, Guss, Nichol, Paino, Tezak, Tang, Babuschkin, Balaji, Jain, Saunders, Hesse, Carr, Leike, Achiam, Misra, Morikawa, Radford, Knight, Brundage, Murati, Mayer, Welinder, McGrew, Amodei, McCandlish, Sutskever, and Zaremba]{chen2021codex}
Mark Chen, Jerry Tworek, Heewoo Jun, Qiming Yuan, Henrique~Ponde de~Oliveira~Pinto, Jared Kaplan, Harri Edwards, Yuri Burda, Nicholas Joseph, Greg Brockman, Alex Ray, Raul Puri, Gretchen Krueger, Michael Petrov, Heidy Khlaaf, Girish Sastry, Pamela Mishkin, Brooke Chan, Scott Gray, Nick Ryder, Mikhail Pavlov, Alethea Power, Lukasz Kaiser, Mohammad Bavarian, Clemens Winter, Philippe Tillet, Felipe~Petroski Such, Dave Cummings, Matthias Plappert, Fotios Chantzis, Elizabeth Barnes, Ariel Herbert-Voss, William~Hebgen Guss, Alex Nichol, Alex Paino, Nikolas Tezak, Jie Tang, Igor Babuschkin, Suchir Balaji, Shantanu Jain, William Saunders, Christopher Hesse, Andrew~N. Carr, Jan Leike, Josh Achiam, Vedant Misra, Evan Morikawa, Alec Radford, Matthew Knight, Miles Brundage, Mira Murati, Katie Mayer, Peter Welinder, Bob McGrew, Dario Amodei, Sam McCandlish, Ilya Sutskever, and Wojciech Zaremba.
\newblock Evaluating large language models trained on code.
\newblock 2021.

\bibitem[Devlin et~al.(2019)Devlin, Chang, Lee, and Toutanova]{devlin-etal-2019-bert}
Jacob Devlin, Ming-Wei Chang, Kenton Lee, and Kristina Toutanova.
\newblock {BERT}: Pre-training of deep bidirectional transformers for language understanding.
\newblock In \emph{Proceedings of the 2019 Conference of the North {A}merican Chapter of the Association for Computational Linguistics: Human Language Technologies, Volume 1 (Long and Short Papers)}, pp.\  4171--4186, Minneapolis, Minnesota, June 2019. Association for Computational Linguistics.
\newblock \doi{10.18653/v1/N19-1423}.
\newblock URL \url{https://aclanthology.org/N19-1423}.

\bibitem[Dubey et~al.(2024)Dubey, Jauhri, Pandey, Kadian, Al-Dahle, Letman, Mathur, Schelten, Yang, Fan, et~al.]{dubey2024llama}
Abhimanyu Dubey, Abhinav Jauhri, Abhinav Pandey, Abhishek Kadian, Ahmad Al-Dahle, Aiesha Letman, Akhil Mathur, Alan Schelten, Amy Yang, Angela Fan, et~al.
\newblock The llama 3 herd of models.
\newblock \emph{arXiv preprint arXiv:2407.21783}, 2024.

\bibitem[Durmus et~al.()Durmus, Tamkin, Clark, Wei, Marcus, Batson, Handa, Lovitt, Tong, McCain, et~al.]{durmusevaluating}
Esin Durmus, Alex Tamkin, Jack Clark, Jerry Wei, Jonathan Marcus, Joshua Batson, Kunal Handa, Liane Lovitt, Meg Tong, Miles McCain, et~al.
\newblock Evaluating feature steering: A case study in mitigating social biases, 2024.
\newblock \emph{URL https://anthropic. com/research/evaluating-feature-steering}.

\bibitem[Elazar et~al.(2021)Elazar, Ravfogel, Jacovi, and Goldberg]{elazar2021amnesic}
Yanai Elazar, Shauli Ravfogel, Alon Jacovi, and Yoav Goldberg.
\newblock Amnesic probing: Behavioral explanation with amnesic counterfactuals.
\newblock \emph{Transactions of the Association for Computational Linguistics}, 9:\penalty0 160--175, 2021.

\bibitem[Florian et~al.(2024)Florian, Alexandre, Benjamin, Yann, and Alexandre]{florian2024exploring}
Le~Bronnec Florian, Verine Alexandre, Negrevergne Benjamin, Chevaleyre Yann, and Allauzen Alexandre.
\newblock Exploring precision and recall to assess the quality and diversity of llms.
\newblock \emph{arXiv preprint arXiv:2402.10693}, 2024.

\bibitem[Hendrycks et~al.(2020)Hendrycks, Burns, Basart, Zou, Mazeika, Song, and Steinhardt]{hendrycks2020measuring}
Dan Hendrycks, Collin Burns, Steven Basart, Andy Zou, Mantas Mazeika, Dawn Song, and Jacob Steinhardt.
\newblock Measuring massive multitask language understanding.
\newblock \emph{arXiv preprint arXiv:2009.03300}, 2020.

\bibitem[Hendrycks et~al.(2021)Hendrycks, Carlini, Schulman, and Steinhardt]{hendrycks2021unsolved}
Dan Hendrycks, Nicholas Carlini, John Schulman, and Jacob Steinhardt.
\newblock Unsolved problems in ml safety.
\newblock \emph{arXiv preprint arXiv:2109.13916}, 2021.

\bibitem[Hutchinson et~al.(2020)Hutchinson, Prabhakaran, Denton, Webster, Zhong, and Denuyl]{hutchinson-etal-2020-social}
Ben Hutchinson, Vinodkumar Prabhakaran, Emily Denton, Kellie Webster, Yu~Zhong, and Stephen Denuyl.
\newblock Social biases in {NLP} models as barriers for persons with disabilities.
\newblock In \emph{Proceedings of the 58th Annual Meeting of the Association for Computational Linguistics}, pp.\  5491--5501, Online, July 2020. Association for Computational Linguistics.
\newblock \doi{10.18653/v1/2020.acl-main.487}.
\newblock URL \url{https://aclanthology.org/2020.acl-main.487}.

\bibitem[Jorgensen et~al.(2023)Jorgensen, Cope, Schoots, and Shanahan]{jorgensen2023improving}
Ole Jorgensen, Dylan Cope, Nandi Schoots, and Murray Shanahan.
\newblock Improving activation steering in language models with mean-centring.
\newblock \emph{arXiv preprint arXiv:2312.03813}, 2023.

\bibitem[Leong et~al.(2023)Leong, Cheng, Wang, Wang, and Li]{leong2023self}
Chak~Tou Leong, Yi~Cheng, Jiashuo Wang, Jian Wang, and Wenjie Li.
\newblock Self-detoxifying language models via toxification reversal.
\newblock \emph{arXiv preprint arXiv:2310.09573}, 2023.

\bibitem[Li et~al.(2024)Li, Zheng, and Huang]{li2024open}
Tianlong Li, Xiaoqing Zheng, and Xuanjing Huang.
\newblock Open the pandora's box of llms: Jailbreaking llms through representation engineering.
\newblock \emph{arXiv preprint arXiv:2401.06824}, 2024.

\bibitem[Lin et~al.(2022)Lin, Hilton, and Evans]{lin-etal-2022-truthfulqa}
Stephanie Lin, Jacob Hilton, and Owain Evans.
\newblock {T}ruthful{QA}: Measuring how models mimic human falsehoods.
\newblock In \emph{Proceedings of the 60th Annual Meeting of the Association for Computational Linguistics (Volume 1: Long Papers)}, pp.\  3214--3252, Dublin, Ireland, May 2022. Association for Computational Linguistics.
\newblock \doi{10.18653/v1/2022.acl-long.229}.
\newblock URL \url{https://aclanthology.org/2022.acl-long.229}.

\bibitem[Liu et~al.(2023)Liu, Wang, Wu, Li, Lv, Ling, Zhu, Zhang, Zheng, and Huang]{liu2023aligning}
Wenhao Liu, Xiaohua Wang, Muling Wu, Tianlong Li, Changze Lv, Zixuan Ling, Jianhao Zhu, Cenyuan Zhang, Xiaoqing Zheng, and Xuanjing Huang.
\newblock Aligning large language models with human preferences through representation engineering.
\newblock \emph{arXiv preprint arXiv:2312.15997}, 2023.

\bibitem[Marks et~al.(2024)Marks, Rager, Michaud, Belinkov, Bau, and Mueller]{marks2024sparse}
Samuel Marks, Can Rager, Eric~J Michaud, Yonatan Belinkov, David Bau, and Aaron Mueller.
\newblock Sparse feature circuits: Discovering and editing interpretable causal graphs in language models.
\newblock \emph{arXiv preprint arXiv:2403.19647}, 2024.

\bibitem[Nadeem et~al.(2020)Nadeem, Bethke, and Reddy]{nadeem2020stereoset}
Moin Nadeem, Anna Bethke, and Siva Reddy.
\newblock Stereoset: Measuring stereotypical bias in pretrained language models, 2020.

\bibitem[Ngo(2022)]{ngo2022alignment}
Richard Ngo.
\newblock The alignment problem from a deep learning perspective.
\newblock \emph{arXiv preprint arXiv:2209.00626}, 2022.

\bibitem[Nori et~al.(2023)Nori, King, McKinney, Carignan, and Horvitz]{nori2023capabilities}
Harsha Nori, Nicholas King, Scott~Mayer McKinney, Dean Carignan, and Eric Horvitz.
\newblock Capabilities of gpt-4 on medical challenge problems.
\newblock \emph{arXiv preprint arXiv:2303.13375}, 2023.

\bibitem[OpenAI(2023)]{gpt4}
OpenAI.
\newblock Gpt-4 technical report, 2023.

\bibitem[Pan et~al.(2022)Pan, Bhatia, and Steinhardt]{pan2022the}
Alexander Pan, Kush Bhatia, and Jacob Steinhardt.
\newblock The effects of reward misspecification: Mapping and mitigating misaligned models.
\newblock In \emph{International Conference on Learning Representations}, 2022.
\newblock URL \url{https://openreview.net/forum?id=JYtwGwIL7ye}.

\bibitem[Park et~al.(2023)Park, O'Brien, Cai, Morris, Liang, and Bernstein]{park2023generative}
Joon~Sung Park, Joseph~C O'Brien, Carrie~J Cai, Meredith~Ringel Morris, Percy Liang, and Michael~S Bernstein.
\newblock Generative agents: Interactive simulacra of human behavior.
\newblock \emph{arXiv preprint arXiv:2304.03442}, 2023.

\bibitem[Radford et~al.(2019)Radford, Wu, Child, Luan, Amodei, and Sutskever]{Radford2019LanguageMA}
Alec Radford, Jeff Wu, Rewon Child, David Luan, Dario Amodei, and Ilya Sutskever.
\newblock Language models are unsupervised multitask learners.
\newblock 2019.

\bibitem[Rae et~al.(2021)Rae, Borgeaud, Cai, Millican, Hoffmann, Song, Aslanides, Henderson, Ring, Young, et~al.]{rae2021scaling}
Jack~W Rae, Sebastian Borgeaud, Trevor Cai, Katie Millican, Jordan Hoffmann, Francis Song, John Aslanides, Sarah Henderson, Roman Ring, Susannah Young, et~al.
\newblock Scaling language models: Methods, analysis \& insights from training gopher.
\newblock \emph{arXiv preprint arXiv:2112.11446}, 2021.

\bibitem[Robey et~al.(2021)Robey, Chamon, Pappas, Hassani, and Ribeiro]{robey2021adversarial}
Alexander Robey, Luiz Chamon, George~J Pappas, Hamed Hassani, and Alejandro Ribeiro.
\newblock Adversarial robustness with semi-infinite constrained learning.
\newblock \emph{Advances in Neural Information Processing Systems}, 34:\penalty0 6198--6215, 2021.

\bibitem[Robey et~al.(2022)Robey, Chamon, Pappas, and Hassani]{robey2022probabilistically}
Alexander Robey, Luiz Chamon, George~J Pappas, and Hamed Hassani.
\newblock Probabilistically robust learning: Balancing average and worst-case performance.
\newblock In \emph{International Conference on Machine Learning}, pp.\  18667--18686. PMLR, 2022.

\bibitem[Roose(2023)]{nytimes_marry_reporter}
Kevin Roose.
\newblock A conversation with bing’s chatbot left me deeply unsettled.
\newblock \emph{New York Times}, 2023.

\bibitem[R{\"o}ttger et~al.(2023)R{\"o}ttger, Kirk, Vidgen, Attanasio, Bianchi, and Hovy]{rottger2023xstest}
Paul R{\"o}ttger, Hannah~Rose Kirk, Bertie Vidgen, Giuseppe Attanasio, Federico Bianchi, and Dirk Hovy.
\newblock Xstest: A test suite for identifying exaggerated safety behaviours in large language models.
\newblock \emph{arXiv preprint arXiv:2308.01263}, 2023.

\bibitem[Schulman et~al.(2023)Schulman, Zoph, Kim, Hilton, Menick, Weng, Felipe, Uribe, Fedus, Metz, Pokorny, Lopes, Zhao, Vijayvergiya, Sigler, Perelman, Voss, Heaton, Parish, Cummings, Nayak, Balcom, Schnurr, Kaftan, Hallacy, Turley, Deutsch, Goel, Ward, Konstantinidis, Zaremba, Ouyang, Bogdonoff, Gross, Medina, Yoo, Lee, Lowe, Mossing, Huizinga, Jiang, Wainwright, Almeida, Lin, Zhang, Xiao, Slama, Bills, Gray, Leike, Pachocki, Tillet, Jain, Brockman, Ryder, Paino, Yuan, Winter, Wang, Bavarian, Babuschkin, Sidor, Kanitscheider, Pavlov, Plappert, Tezak, Jun, Zhuk, Pong, Kaiser, Tworek, Carr, Weng, Agarwal, Cobbe, Kosaraju, Power, Polu, Han, Puri, Jain, Chess, Gibson, Boiko, Parparita, Tootoonchian, Kosic, and Hesse]{chatgpt}
John Schulman, Barret Zoph, Christina Kim, Jacob Hilton, Jacob Menick, Jiayi Weng, Juan Felipe, Ceron Uribe, Liam Fedus, Luke Metz, Michael Pokorny, Rapha~Gontijo Lopes, Shengjia Zhao, Arun Vijayvergiya, Eric Sigler, Adam Perelman, Chelsea Voss, Mike Heaton, Joel Parish, Dave Cummings, Rajeev Nayak, Valerie Balcom, David Schnurr, Tomer Kaftan, Chris Hallacy, Nicholas Turley, Noah Deutsch, Vik Goel, Jonathan Ward, Aris Konstantinidis, Wojciech Zaremba, Long Ouyang, Leonard Bogdonoff, Joshua Gross, David Medina, Sarah Yoo, Teddy Lee, Ryan Lowe, Dan Mossing, Joost Huizinga, Roger Jiang, Carroll Wainwright, Diogo Almeida, Steph Lin, Marvin Zhang, Kai Xiao, Katarina Slama, Steven Bills, Alex Gray, Jan Leike, Jakub Pachocki, Phil Tillet, Shantanu Jain, Greg Brockman, Nick Ryder, Alex Paino, Qiming Yuan, Clemens Winter, Ben Wang, Mo~Bavarian, Igor Babuschkin, Szymon Sidor, Ingmar Kanitscheider, Mikhail Pavlov, Matthias Plappert, Nik Tezak, Heewoo Jun, William Zhuk, Vitchyr Pong, Lukasz Kaiser, Jerry Tworek, Andrew
  Carr, Lilian Weng, Sandhini Agarwal, Karl Cobbe, Vineet Kosaraju, Alethea Power, Stanislas Polu, Jesse Han, Raul Puri, Shawn Jain, Benjamin Chess, Christian Gibson, Oleg Boiko, Emy Parparita, Amin Tootoonchian, Kyle Kosic, and Christopher Hesse.
\newblock Introducing chatgpt.
\newblock \emph{OpenAI blog}, 2023.

\bibitem[Shalev-Shwartz et~al.(2020)Shalev-Shwartz, Shammah, and Shashua]{shalev2020ethics}
Shai Shalev-Shwartz, Shaked Shammah, and Amnon Shashua.
\newblock On the ethics of building ai in a responsible manner.
\newblock \emph{arXiv preprint arXiv:2004.04644}, 2020.

\bibitem[Subhash(2023)]{subhash2023can}
Varshini Subhash.
\newblock Can large language models change user preference adversarially?
\newblock \emph{arXiv preprint arXiv:2302.10291}, 2023.

\bibitem[Tan et~al.(2024)Tan, li, Zhu, Bu, Su, Yue, and Bo]{tan2024truly}
Yingshui Tan, Yanshi li, Xiaoyong Zhu, Xingyuan Bu, Wenbo Su, Xiangyu Yue, and Zheng Bo.
\newblock Truly safe \& truly helpful: Achieving harmonious balance for large language model.
\newblock \emph{Openreview}, 2024.

\bibitem[Taylor et~al.(2016)Taylor, Yudkowsky, LaVictoire, and Critch]{taylor2016alignment}
Jessica Taylor, Eliezer Yudkowsky, Patrick LaVictoire, and Andrew Critch.
\newblock Alignment for advanced machine learning systems.
\newblock \emph{Ethics of Artificial Intelligence}, pp.\  342--382, 2016.

\bibitem[Templeton(2024)]{templeton2024scaling}
Adly Templeton.
\newblock \emph{Scaling monosemanticity: Extracting interpretable features from claude 3 sonnet}.
\newblock Anthropic, 2024.

\bibitem[Touvron et~al.(2023)Touvron, Martin, Stone, Albert, Almahairi, Babaei, Bashlykov, Batra, Bhargava, Bhosale, et~al.]{touvron2023llama}
Hugo Touvron, Louis Martin, Kevin Stone, Peter Albert, Amjad Almahairi, Yasmine Babaei, Nikolay Bashlykov, Soumya Batra, Prajjwal Bhargava, Shruti Bhosale, et~al.
\newblock Llama 2: Open foundation and fine-tuned chat models.
\newblock \emph{arXiv preprint arXiv:2307.09288}, 2023.

\bibitem[Turner et~al.(2023)Turner, Thiergart, Udell, Leech, Mini, and MacDiarmid]{turner2023activation}
Alex Turner, Lisa Thiergart, David Udell, Gavin Leech, Ulisse Mini, and Monte MacDiarmid.
\newblock Activation addition: Steering language models without optimization.
\newblock \emph{arXiv preprint arXiv:2308.10248}, 2023.

\bibitem[van~der Weij et~al.(2024)van~der Weij, Poesio, and Schoots]{van2024extending}
Teun van~der Weij, Massimo Poesio, and Nandi Schoots.
\newblock Extending activation steering to broad skills and multiple behaviours.
\newblock \emph{arXiv preprint arXiv:2403.05767}, 2024.

\bibitem[Venkit et~al.(2022)Venkit, Srinath, and Wilson]{venkit-etal-2022-study}
Pranav~Narayanan Venkit, Mukund Srinath, and Shomir Wilson.
\newblock A study of implicit bias in pretrained language models against people with disabilities.
\newblock In \emph{Proceedings of the 29th International Conference on Computational Linguistics}, pp.\  1324--1332, Gyeongju, Republic of Korea, October 2022. International Committee on Computational Linguistics.
\newblock URL \url{https://aclanthology.org/2022.coling-1.113}.

\bibitem[Wallace et~al.(2019)Wallace, Feng, Kandpal, Gardner, and Singh]{wallace-etal-2019-universal}
Eric Wallace, Shi Feng, Nikhil Kandpal, Matt Gardner, and Sameer Singh.
\newblock Universal adversarial triggers for attacking and analyzing {NLP}.
\newblock In \emph{Proceedings of the 2019 Conference on Empirical Methods in Natural Language Processing and the 9th International Joint Conference on Natural Language Processing (EMNLP-IJCNLP)}, pp.\  2153--2162, Hong Kong, China, November 2019. Association for Computational Linguistics.
\newblock \doi{10.18653/v1/D19-1221}.
\newblock URL \url{https://aclanthology.org/D19-1221}.

\bibitem[Wang et~al.(2024{\natexlab{a}})Wang, Zhang, Xu, Xi, Deng, Yao, Zhang, Yang, Wang, and Chen]{wang2024detoxifying}
Mengru Wang, Ningyu Zhang, Ziwen Xu, Zekun Xi, Shumin Deng, Yunzhi Yao, Qishen Zhang, Linyi Yang, Jindong Wang, and Huajun Chen.
\newblock Detoxifying large language models via knowledge editing.
\newblock \emph{arXiv preprint arXiv:2403.14472}, 2024{\natexlab{a}}.

\bibitem[Wang et~al.(2024{\natexlab{b}})Wang, Zhang, Li, Tan, Wang, Ren, Jiang, and Qiu]{wang2024inferaligner}
Pengyu Wang, Dong Zhang, Linyang Li, Chenkun Tan, Xinghao Wang, Ke~Ren, Botian Jiang, and Xipeng Qiu.
\newblock Inferaligner: Inference-time alignment for harmlessness through cross-model guidance.
\newblock \emph{arXiv preprint arXiv:2401.11206}, 2024{\natexlab{b}}.

\bibitem[Wei et~al.(2024)Wei, Huang, Huang, Xie, Qi, Xia, Mittal, Wang, and Henderson]{wei2024assessing}
Boyi Wei, Kaixuan Huang, Yangsibo Huang, Tinghao Xie, Xiangyu Qi, Mengzhou Xia, Prateek Mittal, Mengdi Wang, and Peter Henderson.
\newblock Assessing the brittleness of safety alignment via pruning and low-rank modifications.
\newblock \emph{arXiv preprint arXiv:2402.05162}, 2024.

\bibitem[Weidinger et~al.(2022)Weidinger, Uesato, Rauh, Griffin, Huang, Mellor, Glaese, Cheng, Balle, Kasirzadeh, Biles, Brown, Kenton, Hawkins, Stepleton, Birhane, Hendricks, Rimell, Isaac, Haas, Legassick, Irving, and Gabriel]{weidinger2022taxonomy}
Laura Weidinger, Jonathan Uesato, Maribeth Rauh, Conor Griffin, Po-Sen Huang, John Mellor, Amelia Glaese, Myra Cheng, Borja Balle, Atoosa Kasirzadeh, Courtney Biles, Sasha Brown, Zac Kenton, Will Hawkins, Tom Stepleton, Abeba Birhane, Lisa~Anne Hendricks, Laura Rimell, William Isaac, Julia Haas, Sean Legassick, Geoffrey Irving, and Iason Gabriel.
\newblock Taxonomy of risks posed by language models.
\newblock In \emph{2022 ACM Conference on Fairness, Accountability, and Transparency}, FAccT '22, pp.\  214–229, New York, NY, USA, 2022. Association for Computing Machinery.
\newblock ISBN 9781450393522.
\newblock \doi{10.1145/3531146.3533088}.
\newblock URL \url{https://doi.org/10.1145/3531146.3533088}.

\bibitem[West(2023)]{west2023advances}
Colin~G West.
\newblock Advances in apparent conceptual physics reasoning in gpt-4.
\newblock \emph{arXiv e-prints}, pp.\  arXiv--2303, 2023.

\bibitem[Wolf et~al.(2023)Wolf, Wies, Levine, and Shashua]{wolf2023fundamental}
Yotam Wolf, Noam Wies, Yoav Levine, and Amnon Shashua.
\newblock Fundamental limitations of alignment in large language models.
\newblock \emph{arXiv preprint arXiv:2304.11082}, 2023.

\bibitem[Wu et~al.(2024)Wu, Liu, Wang, Li, Lv, Ling, Zhu, Zhang, Zheng, and Huang]{wu2024advancing}
Muling Wu, Wenhao Liu, Xiaohua Wang, Tianlong Li, Changze Lv, Zixuan Ling, Jianhao Zhu, Cenyuan Zhang, Xiaoqing Zheng, and Xuanjing Huang.
\newblock Advancing parameter efficiency in fine-tuning via representation editing.
\newblock \emph{arXiv preprint arXiv:2402.15179}, 2024.

\bibitem[Xu et~al.(2021)Xu, Ju, Li, Boureau, Weston, and Dinan]{xu-etal-2021-bot}
Jing Xu, Da~Ju, Margaret Li, Y-Lan Boureau, Jason Weston, and Emily Dinan.
\newblock Bot-adversarial dialogue for safe conversational agents.
\newblock In \emph{Proceedings of the 2021 Conference of the North American Chapter of the Association for Computational Linguistics: Human Language Technologies}, pp.\  2950--2968, Online, June 2021. Association for Computational Linguistics.
\newblock \doi{10.18653/v1/2021.naacl-main.235}.
\newblock URL \url{https://aclanthology.org/2021.naacl-main.235}.

\bibitem[Xu et~al.(2024)Xu, Huang, Wang, Wu, Yao, and Xie]{xu2024uncovering}
Zhihao Xu, Ruixuan Huang, Xiting Wang, Fangzhao Wu, Jing Yao, and Xing Xie.
\newblock Uncovering safety risks in open-source llms through concept activation vector.
\newblock \emph{arXiv preprint arXiv:2404.12038}, 2024.

\bibitem[Yan et~al.(2024)Yan, Wang, Li, and Zhang]{yan2024potential}
Jianhao Yan, Futing Wang, Yafu Li, and Yue Zhang.
\newblock Potential and challenges of model editing for social debiasing.
\newblock \emph{arXiv preprint arXiv:2402.13462}, 2024.

\bibitem[Yu \& Sagae(2021)Yu and Sagae]{yu-sagae-2021-automatically}
Dian Yu and Kenji Sagae.
\newblock Automatically exposing problems with neural dialog models.
\newblock In \emph{Proceedings of the 2021 Conference on Empirical Methods in Natural Language Processing}, pp.\  456--470, Online and Punta Cana, Dominican Republic, November 2021. Association for Computational Linguistics.
\newblock \doi{10.18653/v1/2021.emnlp-main.37}.
\newblock URL \url{https://aclanthology.org/2021.emnlp-main.37}.

\bibitem[Yudkowsky(2001)]{yudkowsky2001creating}
Eliezer Yudkowsky.
\newblock Creating friendly ai 1.0: The analysis and design of benevolent goal architectures.
\newblock \emph{The Singularity Institute, San Francisco, USA}, 2001.

\bibitem[Zhang et~al.(2024)Zhang, Yu, and Feng]{zhang2024truthx}
Shaolei Zhang, Tian Yu, and Yang Feng.
\newblock Truthx: Alleviating hallucinations by editing large language models in truthful space.
\newblock \emph{arXiv preprint arXiv:2402.17811}, 2024.

\bibitem[Zou et~al.(2023{\natexlab{a}})Zou, Phan, Chen, Campbell, Guo, Ren, Pan, Yin, Mazeika, Dombrowski, et~al.]{zou2023representation}
Andy Zou, Long Phan, Sarah Chen, James Campbell, Phillip Guo, Richard Ren, Alexander Pan, Xuwang Yin, Mantas Mazeika, Ann-Kathrin Dombrowski, et~al.
\newblock Representation engineering: A top-down approach to ai transparency.
\newblock \emph{arXiv preprint arXiv:2310.01405}, 2023{\natexlab{a}}.

\bibitem[Zou et~al.(2023{\natexlab{b}})Zou, Wang, Kolter, and Fredrikson]{zou2023universal}
Andy Zou, Zifan Wang, J~Zico Kolter, and Matt Fredrikson.
\newblock Universal and transferable adversarial attacks on aligned language models.
\newblock \emph{arXiv preprint arXiv:2307.15043}, 2023{\natexlab{b}}.

\end{thebibliography}
\bibliographystyle{arxiv_upload}

\appendix

\clearpage
\section{Assumptions}\label{sec:assumptions}
In \ref{sec:intro_assumptions} we introduce our assumptions used in proving theorems \ref{theorem:1} and \ref{theorem:2}. We discuss them in \ref{sec:discussion_assumptions} and provide experiments to check their validity in \ref{sec:exp_assumptions}

\subsection{Introduction of assumptions}\label{sec:intro_assumptions}
We assume that for small coefficients of representation steering $r_e$, the norm of the change to the last hidden layer representation is linear in $r_e$:
\begin{assumption}\label{assumption:linear}
    Let $P_{\theta,r_e}(\cdot |q)$ be a language model prompted with query $q$. The change to the last hidden layer representation due to steering with coefficient $r_e$, denoted by $\delta r_e(q) = r^{(L)}(q,r_e) - r^{(L)}(q,0)$ satisfies:
    \begin{equation}
        |\delta r_e(q)| = \lambda |r_e|
    \end{equation}
    For a constant $\lambda >0$ that is query dependent.
\end{assumption}
It is used in theorems \ref{theorem:1} and \ref{theorem:2}, to relate the change to the last hidden layer to the coefficients of injected representations.

A representation of an answer to a query is defined as the latent space embedding of the answer's token, $U^Te_{token}$, where $e_i$ is the one-hot vector of the token $i$ and $U$ is the matrix from the last layer's hidden dimension to the vocabulary. We assume that the representations of positive and negative answers to a query are linearly separable, and that the change to the last hidden layer of the model due to representation engineering linearly classifies them with margin $\Delta$:
\begin{assumption}\label{def:representation_separable}
    Given a query $q$, the change to the last hidden layer of a model due to steering, $\delta r_e(q) = r^{(L)}(q,r_e) - r^{(L)}(q,0)$ , linearly classifies the representations of positive and negative answers to a query $q$ with margin $\Delta$, where the positive and negative answers are defined with respect to a behavior scoring function $B:\Sigma\rightarrow\{-1,+1\}$:
    \begin{equation}
        min_{i: B(i)>0,j: B(j)<0}\bigg\{\big\langle \frac{\delta r_e(q)}{|\delta r_e(q)|},U^Te_i-U^Te_j\big\rangle\bigg\} > \Delta
    \end{equation}
\end{assumption}
That is to say, that on the axis defined by $\delta r_e(q)$, positive and negative representations can be separated, and the minimal distance between representations of positive and negative answers on it is $\Delta$. It is used in theorm \ref{theorem:1}, to obtain that the probability of the aligned answers increases \wrt the misaligned answers as the coefficients of the injected representations increases.

Note that the above assumption can be relaxed from a hard margin to a soft margin assumption, where $\delta r_e(q)$ classifies the representations of positive and negative answers, but part of the misaligned/aligned answers' representations are misclassified as aligned/misaligned. This yields similar results to theorem \ref{theorem:1} that are shown in appendix \ref{sec:soft_margin}.

For queries whose topic is unrelated to the behavior with respect to which steering is performed, we expect the change to the last layer representation to be somewhat random on the highest probability tokens as they answer a question that is unrelated to the behavior whose vectors are injected to the model.
Intuitively, the change to the final layer representation has no preference for a correct token over an incorrect token, so an incorrect answer is just as likely to be on one side or the other of the plane defined by the vertical $\delta r_e (q)$ that passes through the correct answer representation. 

\begin{assumption}\label{assumption:normal}
When sampling an answer to a query $q$ that is unrelated to the behavior of steering, the vector $\delta r_e(q)=r^{(L)}(q,r_e) - r^{(L)}(q,0)$, i.e., the resulting change to the last hidden layer representation due to the steering vectors from all layers, is random with the following coordinate-wise distribution on the $T$ highest probability tokens making $1-\epsilon$ of the probability mass:
    \begin{equation}\label{eq:normal}
        \langle \frac{\delta r_e(q)}{|\delta r_e(q)|},U^Te_i \rangle\sim D
    \end{equation}
    Where $D$ is some continuous distribution with variance $\sigma^2>0$.
\end{assumption}
This defines a random directionality of $\delta r_e(q)$ \wrt~the representations of answers.
It is used in theorem \ref{theorem:2} to formalize that steering is a ``perpendicular" direction to the query's relevant answer representations.

\subsection{Discussion of assumptions}\label{sec:discussion_assumptions}

\textbf{Linear last hidden layer change} (assumption \ref{assumption:linear}): Intuitively, when adding vectors of relatively small norms to each layer, the first order Taylor expansion with respect to the vectors is good, and it scales linearly with the coefficients of the vectors. We observe experimentally in subsection \ref{sec:exp_linear} that for small coefficients, the change is indeed approximately linear. Note that it suffices to assume $|\delta r_e(q)|$ grows monotonically with $|r_e|$, but for simplicity and due to experimental observations we assume the linear dependence.

\textbf{Linear classification with margin $\Delta$} (assumption \ref{def:representation_separable}): We expect the representation engineered vectors $r_e$ to be good classifiers because they are obtained by methods of finding directions in the latent space that maximize the distance between representations of positive and negative textual statements. For example, in \cite{zou2023representation} the first principle component is used as a steering vector, obtained via $pca_1 = argmax_{v}\{\expectation_{good,bad}[|\langle v,r_{good}-r_{bad}\rangle|^2]\}$ and in \cite{jorgensen2023improving} the steering vector is obtained as the average of difference between positive and negative statements $\frac{1}{N}\sum_{i=1}^N(r^i_{good}-r^i_{bad})$. In these examples, $r_{good}$ and $r_{bad}$ are representations of queries and not the latent space embedding of the answers,
as in the definition of $\Delta$-representation-separability, but we expect the steering vectors to behave similarly on them.  
In subsection \ref{sec:exp_separability}, we show that indeed $\delta r_e(q)$ clusters positive and negative responses to harmful queries in the model's latent space. In appendix \ref{sec:soft_margin} we also formulate a theorem equivalent to theorem \ref{theorem:1}, but with an imperfect classifier.

\textbf{Random directionality of last hidden layer change} (assumption \ref{assumption:normal}): 
When answering queries that are unrelated to the behavior being enhanced by steering, the directionality of the injected vectors are expected to be random \wrt~the representations of the answers to the query. Therefore, the highest probability tokens are expected to be injected with random noise. We validate this in the next subsection, by looking at the noise injected into the top 10 highest probability tokens in knowledge queries (which typically make over 90\% of the probability mass).

\clearpage
\subsection{Experiments for assumptions}\label{sec:exp_assumptions}
Here we empirically check the validity of our assumptions and empirically estimate the values of the parameters in the bounds. The experiments were performed on Llama 2 13B and Llama 2 13B chat.
We first verify a linear relation between the steering vector coefficient $r_e$ to the last hidden layer change of assumption \ref{assumption:linear}, which yields $\lambda$. Then, we verify the normal distribution assumption \ref{assumption:normal} and the linear classification of assumption \ref{def:representation_separable}.

\begin{figure}[h!]
    \centering
    \includegraphics[scale=0.6]{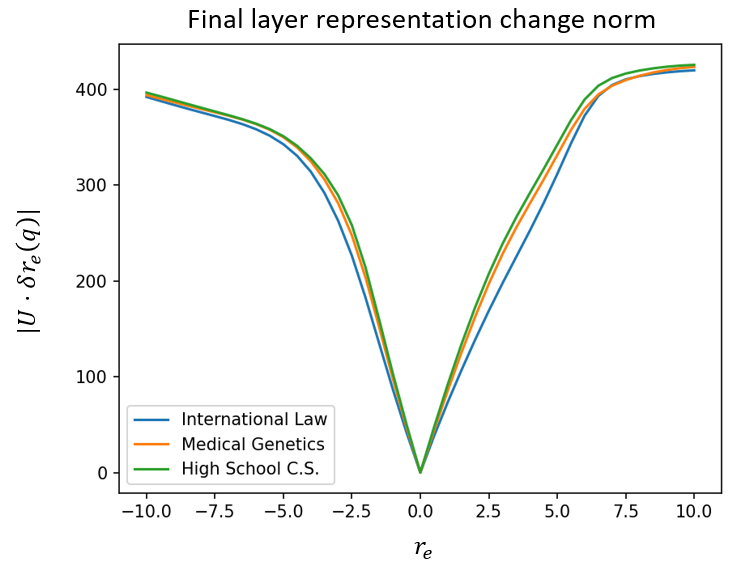}
    \caption{Linear increase in the norm of $U\delta r_e(q)$ for small coefficients, when injected with ``racist" vectors.}
    \label{fig:linear_norm}
\end{figure}

\paragraph{Norm of final hidden layer change is linear in injected vectors}\label{sec:exp_linear}
For a query $q$ we define $\delta r_e(q)=r^{(L)}(q,r_e)-r^{(L)}(q,0)$ as the change of the representation of the query in the final layer. where $r^{(L)}(q,0)$ is the representation if we injected no vector (the default model representation) and $r^{(L)}(q,r_e)$ is the representation given that we inject a vector of norm $r_e$ at a range of layers.
We show that the norm of $\delta r_e(q)$ increases linearly with $r_e$ when $r_e$ is not too large (figure \ref{fig:linear_norm}). Here we use the above mentioned fairness PCA vectors. We average on different queries from a few datasets taken from MMLU. 

In practice we look at $U \delta r_e(q)$, where $U$ is the transformation taking from the final layer representation to the logits vector. Since this is a linear transformation, showing a linear relationship between $r_e$ and $|U \delta r_e(q)|$ implies a linear relationship between $r_e$ and $|\delta r_e(q)|$.

\begin{figure}[h!]
    \centering
    \includegraphics[scale=0.5]{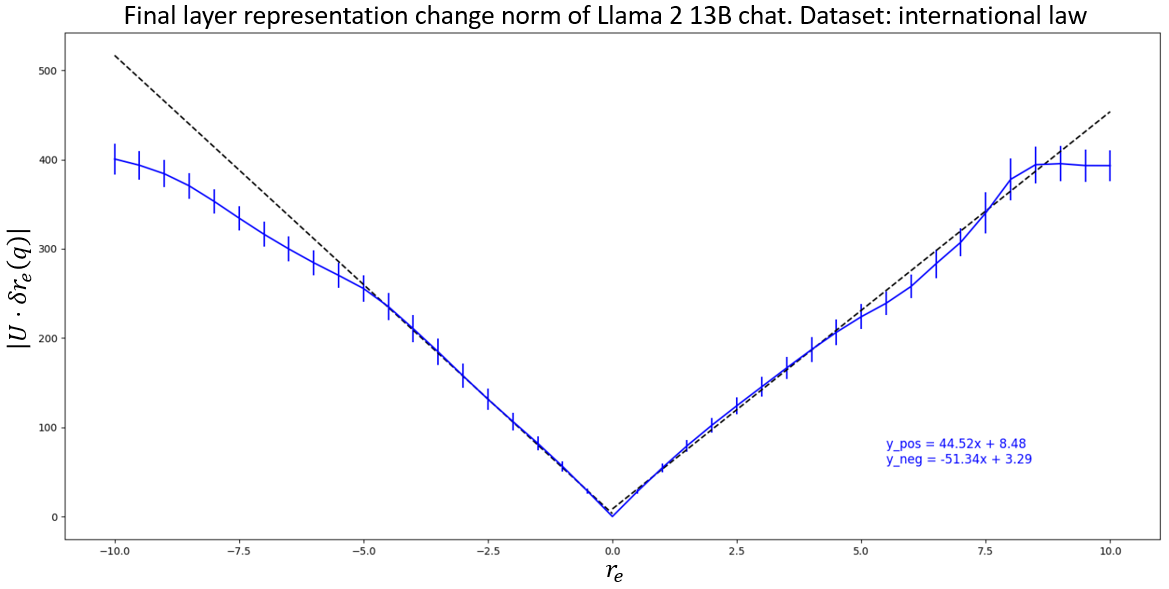}
    \includegraphics[scale=0.5]{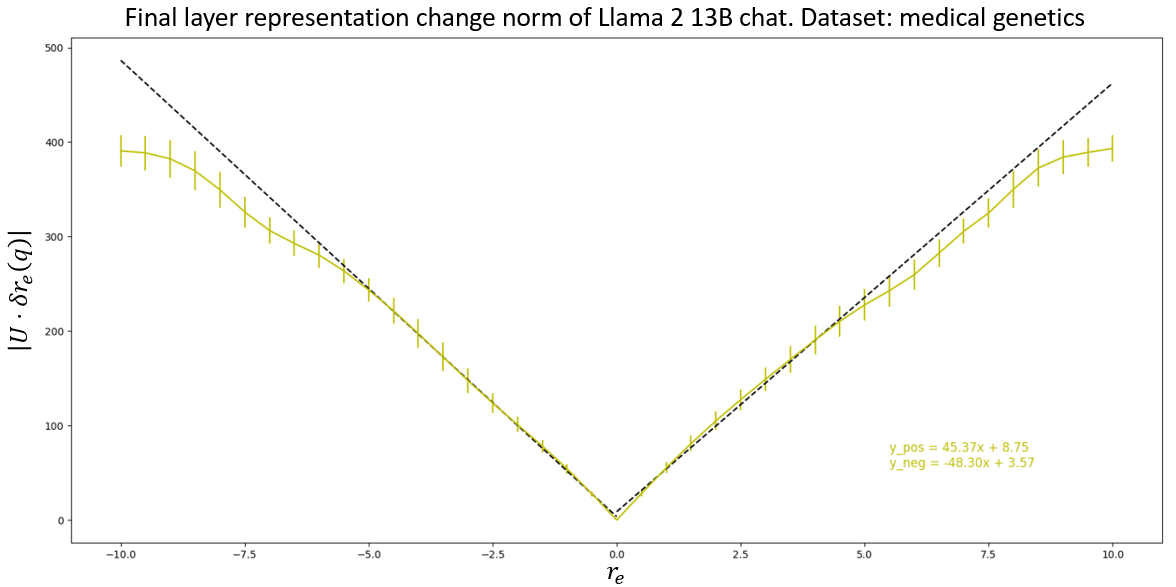}
    \includegraphics[scale=0.5]{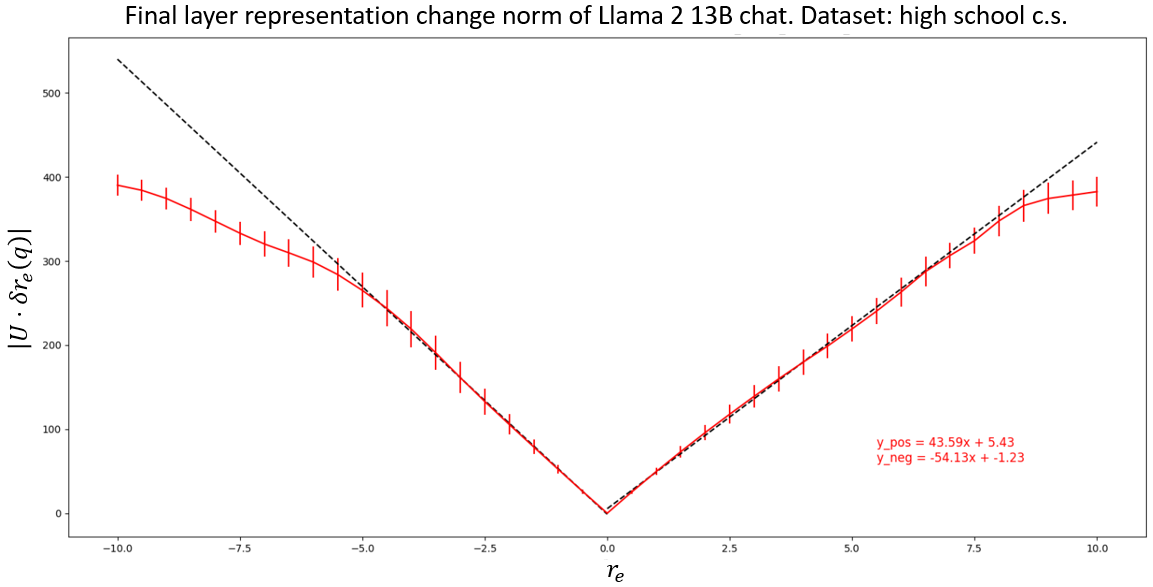}
    \caption{Norm of the final hidden layer representation change as a function of representation engineering coefficient, for Llama 2 13B chat, on different MMLU datasets. The fitted linear curves estimate $\lambda$.}
    \label{fig:linear_fitted_chat}
\end{figure}

\begin{figure}[h!]
    \centering
    \includegraphics[scale=0.5]{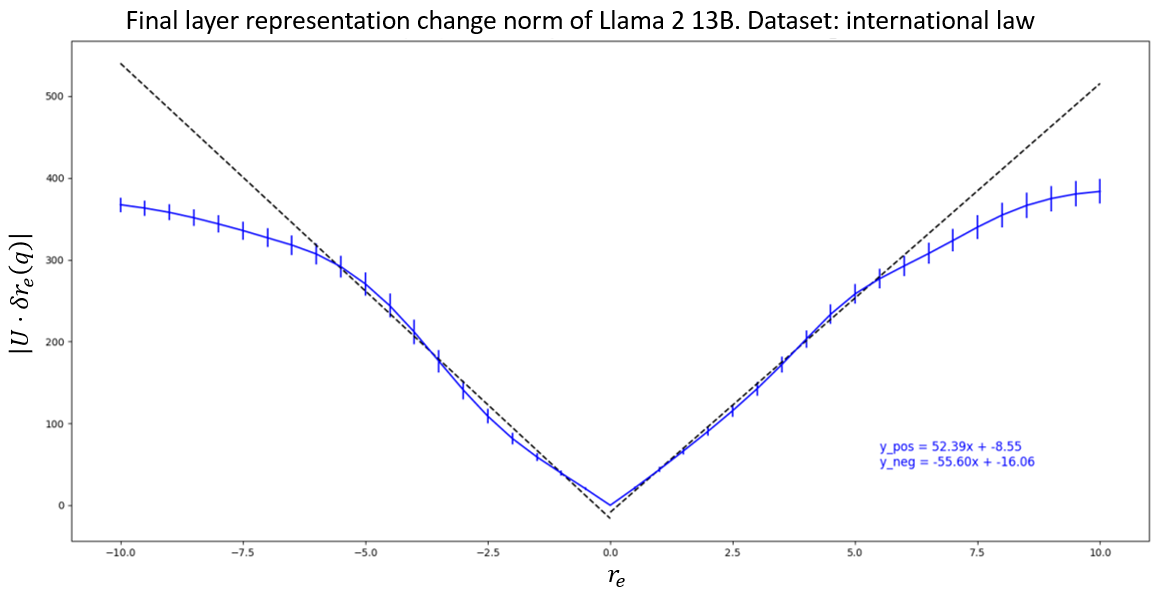}
    \includegraphics[scale=0.5]{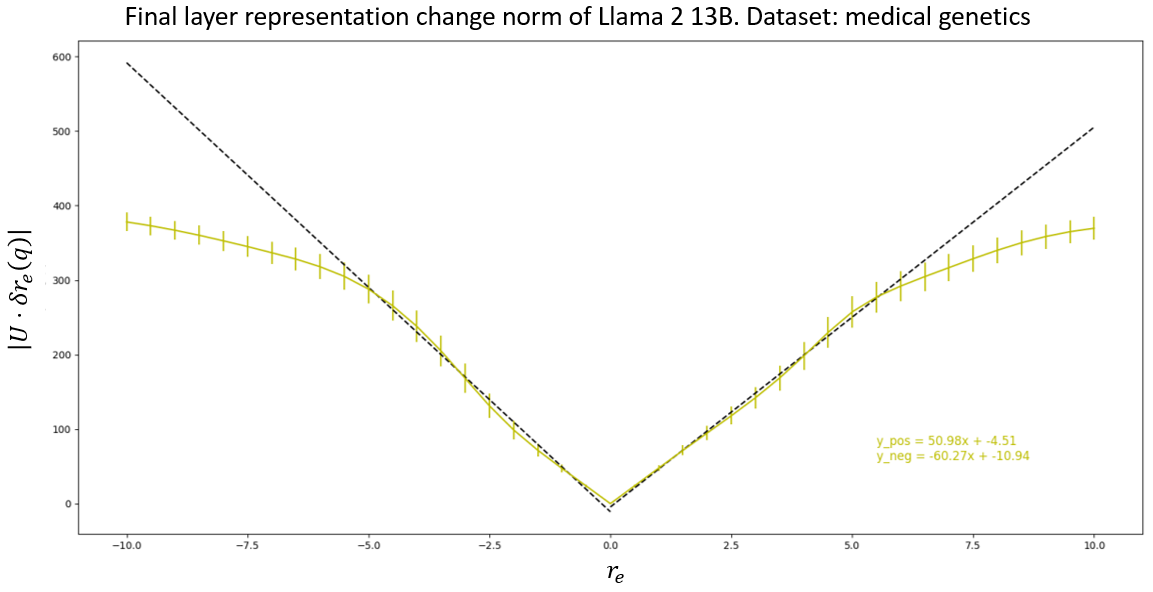}
    \includegraphics[scale=0.5]{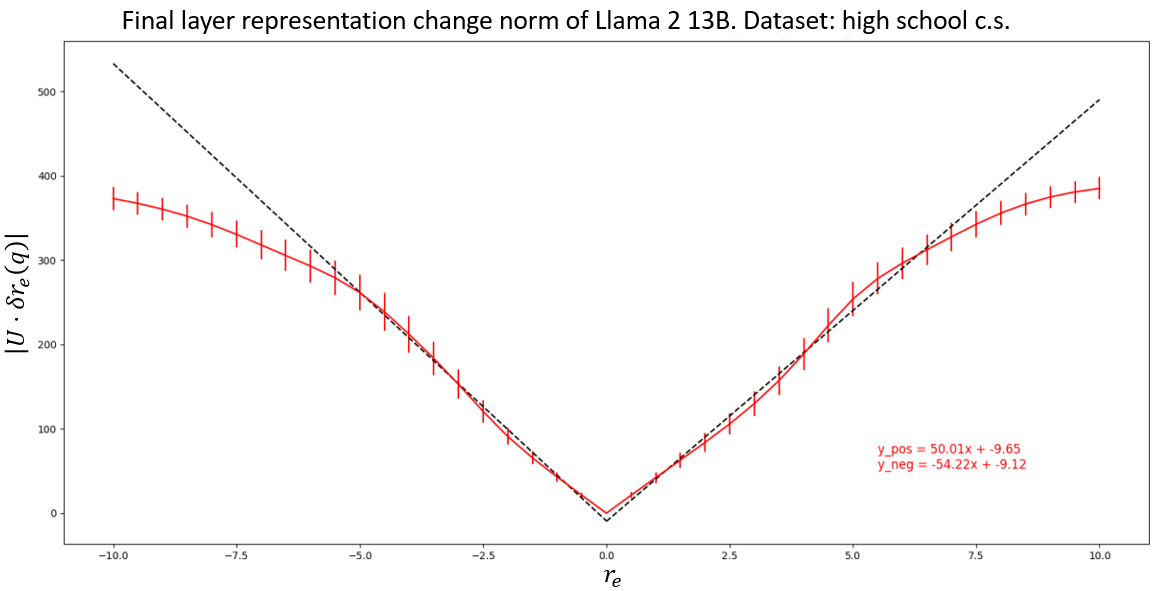}
    \caption{Norm of the final hidden layer representation change as a function of representation engineering coefficient, for Llama 2 13B, on different MMLU datasets. The fitted linear curves estimate $\lambda$.}
    \label{fig:linear_fitted_pretrained}
\end{figure}

In figures \ref{fig:linear_fitted_chat} and \ref{fig:linear_fitted_pretrained} we plot the change in norm for Llama 2 13B chat (injected with racist vectors) and Llama 2 13B (injected with not racist vectors) respectively, on the datasets ``international law", ``medical genetics" and ``high school computer science". We add fitted curves to estimate $\lambda$. We find that it is in the range $40-60$.

\clearpage
\paragraph{Random logit noise assumption}\label{sec:exp_normal}
As proposed in assumption \ref{assumption:normal}, we show here that the projection of a given answer on the representation change $\delta r_e(q)$ is random. (Assuming the question asked is not connected to the property we are changing with the representation engineering). 
In assumption \ref{assumption:normal} we looked at the normalized change: $\langle\frac{\delta r_e(q)}{\lvert\lvert \delta r_e(q) \rvert\rvert},U^Te_i\rangle$. 
Here we will look at $\langle\delta r_e(q),U^Te_i\rangle$, so we expect the distribution to be:
$$\langle\delta r_e(q),U^Te_i\rangle \sim\lvert\lvert \delta r_e(q) \rvert\rvert\cdot D$$
Meaning the standard deviation scales linearly with the norm of $\delta r_e(q)$. 
Since $r_e$ scales linearly with $\delta r_e(q)$, we expect the standard deviation to also scale linearly with $r_e$. 
To measure the effective randomness, we look at $\langle\delta r_e(q),U^T(e_i-e_{correct})\rangle$, which shows explicitly that the correct answer logit change is sometimes enhanced and sometimes decreased relatively to the incorrect answers. We will observe that the noise is approximately normal.

To create the plot, for each question in a dataset, we look at the top 10 answers $e_i,i\in[10]$ (with no representation engineering).
We note that experimentally, the top 10 tokens make the majority of the probability mass (over 90\%). Now for a given $r_e$ coefficient, we calculate the projection of these answers on $\delta r_e(q)$. 
We then aggregate these projections for all the questions in a few dataset and look at their histogram and at their standard deviation. 
We repeat this for different $r_e$ norms. 
\begin{figure}[h]
    \centering
    \includegraphics[scale=0.5]{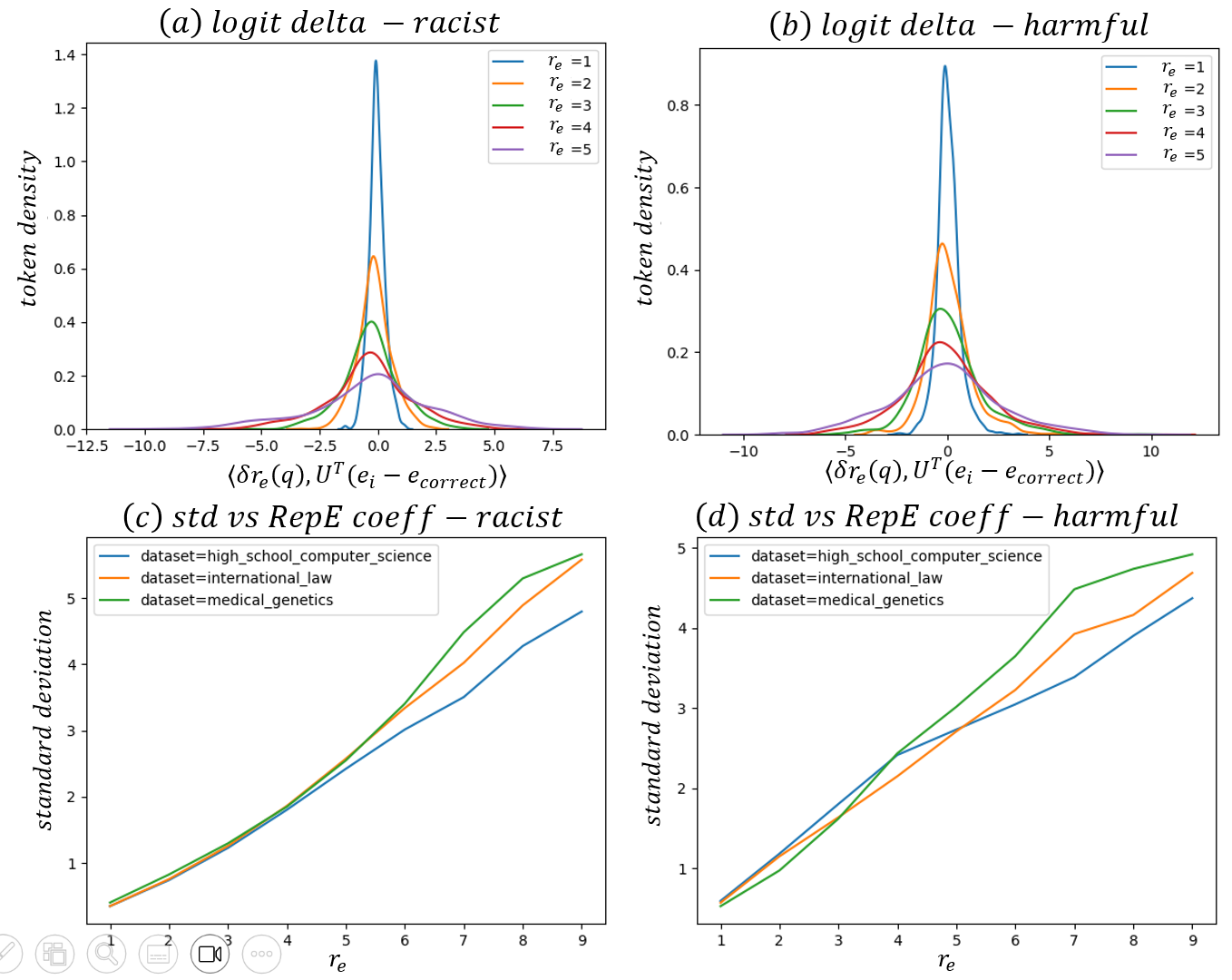}
    \caption{(a) ((b)) Distribution of the change in token logits minus the logit of the correct answer of Llama 2 13B chat when injected with racist (harmful) vectors. As can be seen, it is approximately normal, and in (c) and (d) the standard deviation grows linearly with the coefficient size $r_e$, which is linear in $|\delta r_e(q)|$.}
    \label{fig:logits_normal}
\end{figure}

\begin{figure}[h]
    \centering
    \includegraphics[scale=0.5]{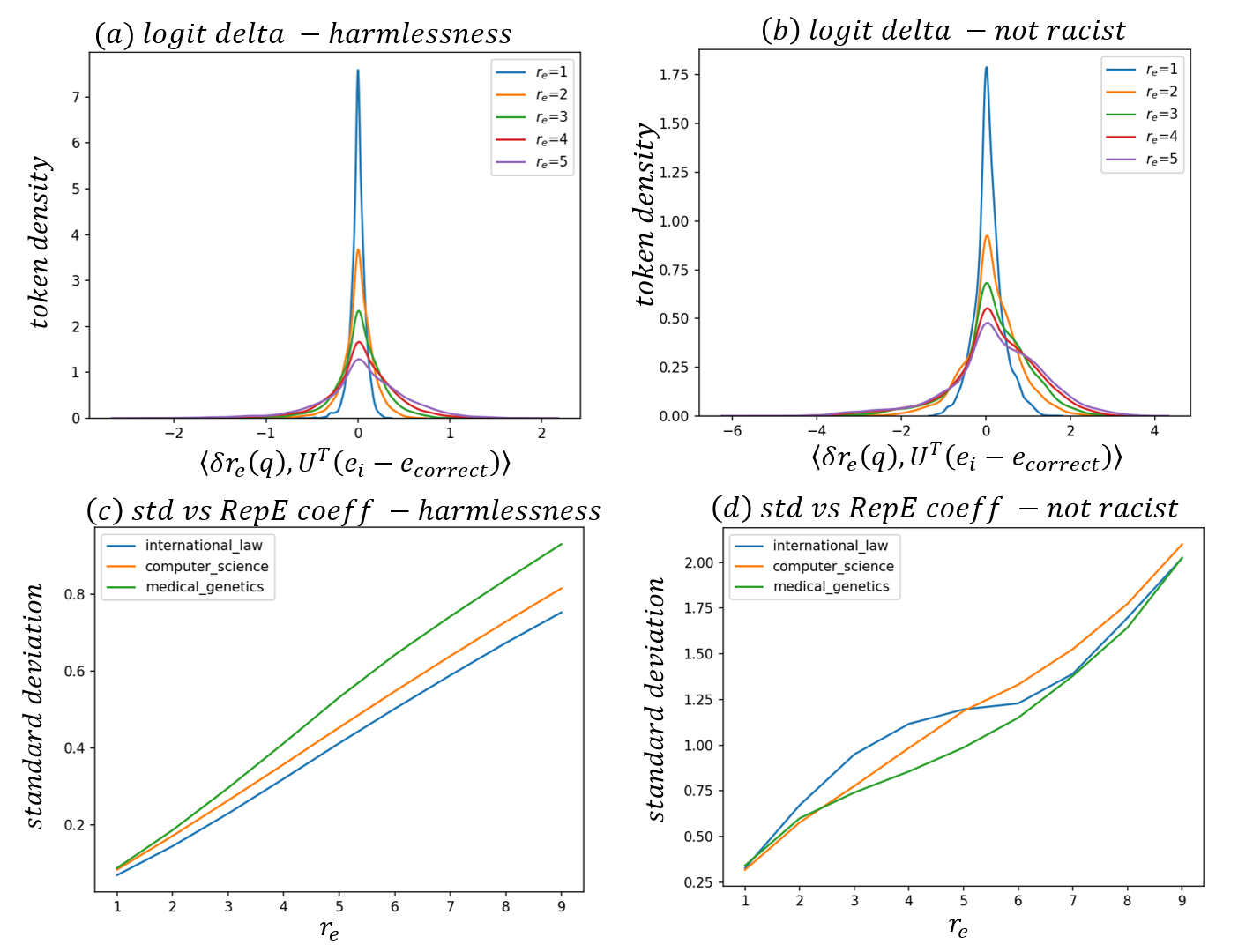}
    \caption{(a) ((b)) Distribution of the change in token logits minus the logit of the correct answer of Llama 2 13B chat when injected with harmless (not-racist) vectors. As can be seen, it is approximately normal, and in (c) and (d) the standard deviation grows linearly with the coefficient size $r_e$, which is linear in $|\delta r_e(q)|$.}
    \label{fig:logits_normal_pretrained}
\end{figure}

The tangent of the curve of figure \ref{fig:logits_normal}c,d is $\lambda \sigma$, as the curve is the standard deviation of $\langle  \frac{\delta r_e(q)}{|\delta r_e(q)|},U^T e_i\rangle\cdot |\delta r_e(q)| = \langle  \frac{\delta r_e(q)}{|\delta r_e(q)|},U^T e_i\rangle\cdot \lambda r_e$, from assumption \ref{assumption:linear}, and the inner product is a random variable of standard deviation $\sigma$, hence the tangent is $\lambda\sigma$. We observe that the noise is approximately normal.
From the linear curve, we estimate $\lambda \sigma = 0.5$, thus $\lambda\sigma\beta\approx 0.8\cdot 0.5$, as it is the mean of a half-normal distribution with parameter $\lambda\sigma$, which is approximately $0.8\lambda\sigma$.

Similarly, for the pretrained model, we find that $\lambda\sigma = 0.2$ and $0.1$ for fairness and harmlessness respectively.

\begin{figure}[h!]
    \centering
    \includegraphics[scale=0.5]{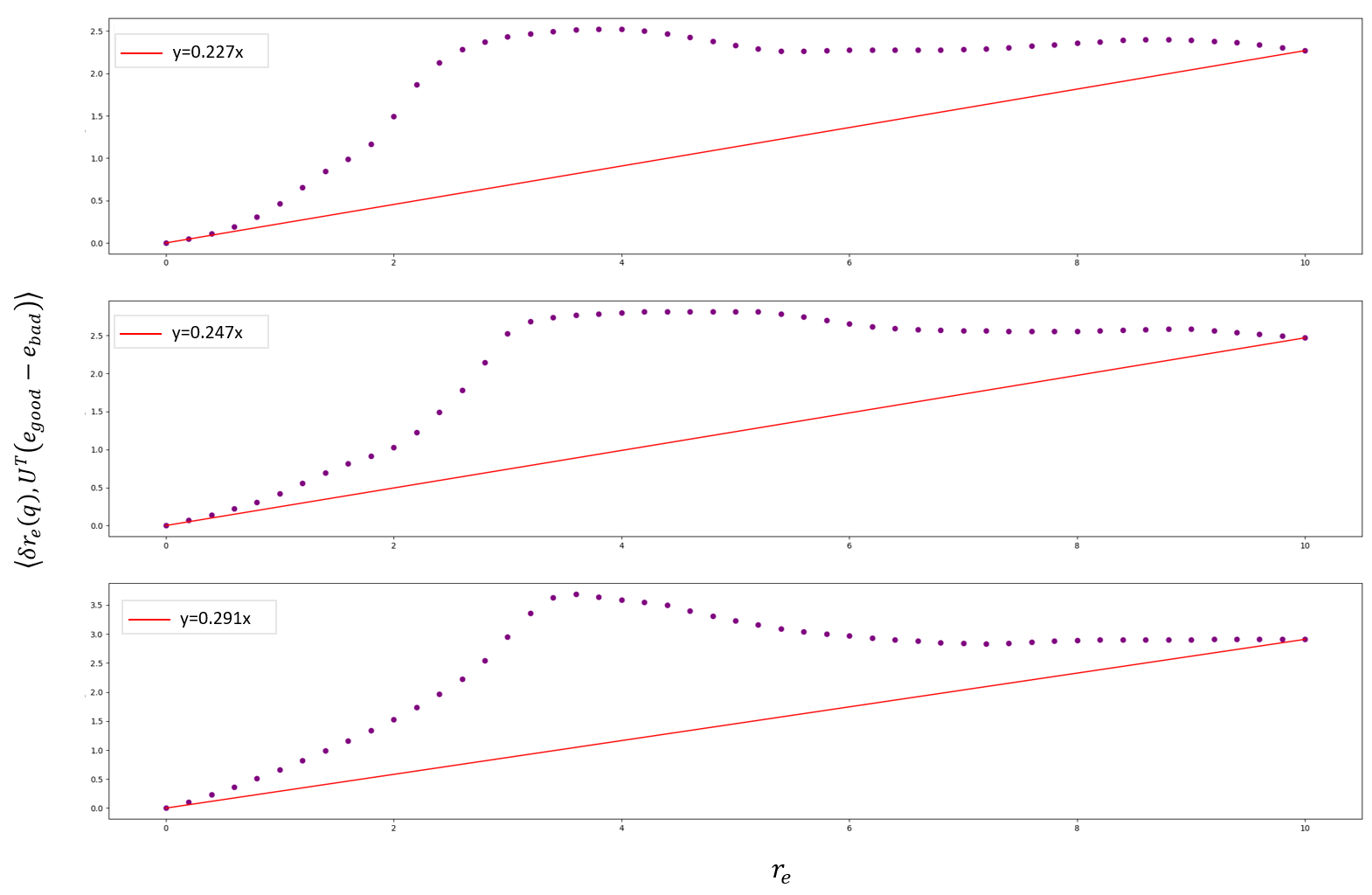}
    \caption{Separation between representation clusters of positive and negative behavior tokens induced by $\delta r_e(q)$ on Llama 2 13B chat for three harmful instructions from the AdvBench dataset.
}
    \label{fig:separation_by_dr_chat}
\end{figure}

\begin{figure}[h!]
    \centering
    \includegraphics[scale=0.5]{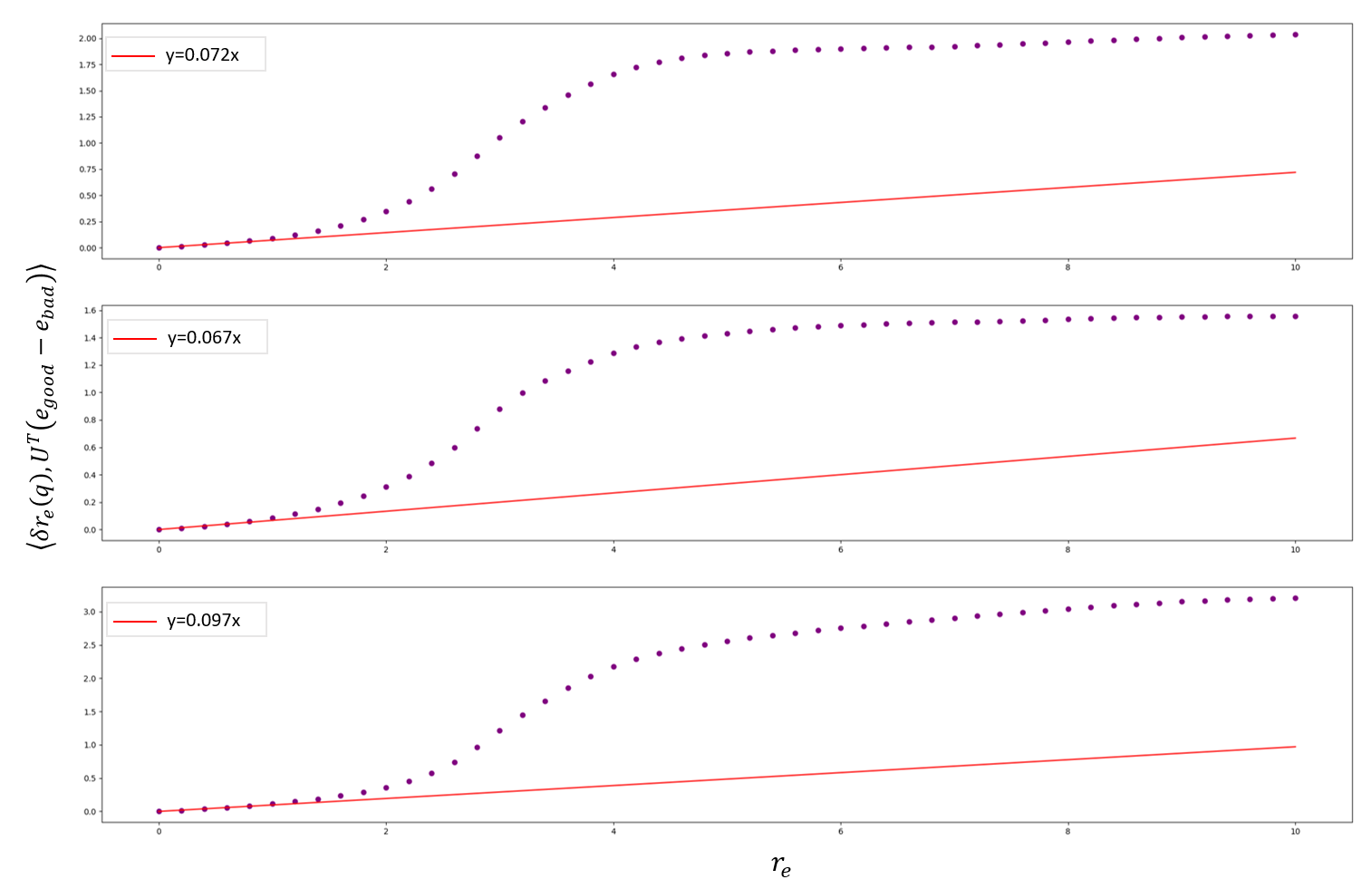}
    \caption{Separation between representation clusters of positive and negative behavior tokens induced by $\delta r_e(q)$ on Llama 2 13B for three harmful instructions from the AdvBench dataset.
}
    \label{fig:separation_by_dr_pretrained}
\end{figure}

\clearpage
\paragraph{Clustering of positive and negative answers to harmful queries}\label{sec:exp_separability}
Here we aim to estimate how well $\Delta$-representation-separability (definition \ref{def:representation_separable}) works in practice. The condition is equivalent to:
\begin{equation}
    \langle \delta r_e(q), U^T(e_{good} - e_{bad})\rangle \geq |\delta r_e(q)|\cdot\Delta
\end{equation}
And by assumption \ref{assumption:linear}, it is equivalent to:
\begin{equation}\label{separation_by_dr}
    \langle \delta r_e(q), U^T(e_{good} - e_{bad})\rangle \geq \Delta\lambda \cdot r_e
\end{equation}
In figure \ref{fig:separation_by_dr_chat} and \ref{fig:separation_by_dr_pretrained}, we plot the distance between the centers of representation clusters for positive and negative answers to harmful queries as the norm of harmful vectors is increased, for Llama 2 13B chat and Llama 2 13B respectively. As can be seen, the distance between the clusters increases, which corresponds to an increase in $\expectation[\langle\delta r_e(q),U^T(e_{good}-e_{bad})\rangle]$. We can define a range of coefficients in which the increase is bounded from below by a linear curve of the form in equation \ref{separation_by_dr}, meaning that the change in the model's representation separates the positive and negative answer representations, similarly to the definition of $\Delta$-representation separability, but with mean instead of min. Thus by equation \ref{separation_by_dr}, the tangent of the lower bounding lines of figures \ref{fig:separation_by_dr_chat} and \ref{fig:separation_by_dr_pretrained} are an estimate for $\Delta\lambda$. From, this we get that $\Delta \lambda$ is approximately $0.1-0.3$. In section \ref{sec:exp}, we obtained values of $\Delta \lambda$ in the rage $0.5-3$ from the free parameter fit on the bound of theorem \ref{theorem:2} to the data.
The difference between these two ranges is attributed to the method of the empirical estimation of $\Delta$ from the linear classification condition that looks for an upper bound on it on the entire $r_e$ range, while the main change in alignment in figure \ref{fig:behavior_expectation} occurs in a more specific range, where the upper bound of $\Delta$ is evidently bigger.

In practice, the good and bad tokens were chosen beforehand as the top 40 tokens of the models when representation engineering is applied and when it is not applied (meaning in one case the model is aligned and in the other it is not).

\clearpage
\section{Proof of theorem 1}\label{proof:theorem_1}
The theorem utilizes assumptions \ref{assumption:linear} and \ref{def:representation_separable}.
The behavior expectation is:
\begin{equation}\label{eq:behavior_expectation}
    B[P_{\theta,r_e}(\cdot|q)] = \frac{\sum_{a_+\in good}P_{\theta,r_e}(a_+|q) - \sum_{a_-\in bad}P_{\theta,r_e}(a_-|q)}{\sum_{a_+\in good}P_{\theta,r_e}(a_+|q) + \sum_{a_-\in bad}P_{\theta,r_e}(a_-|q)} = 
\end{equation}
\begin{equation}
    = \frac{\sum_{a_+\in good}exp(\langle r(q)+\delta r(q),U^Te_{a_+}\rangle) - \sum_{a_-\in bad}exp(\langle r(q)+\delta r(q),U^Te_{a_-} \rangle)}{\sum_{a_+\in good}exp(\langle r(q)+\delta r(q),U^Te_{a_+} \rangle) + \sum_{a_-\in bad}exp(\langle r(q)+\delta r(q),U^Te_{a_-}\rangle)} = 
\end{equation}
Where $r(q)$ is the final hidden layer representation and $\delta r(q)$ is the change to the last hidden layer due to steering on the previous layers. $a_+\in good$ and $a_-\in bad$ denote the aligned and misaligned answers respectively, \ie~ $B(a_\pm) = \pm 1$.
\begin{equation}
    = \frac{1 - \frac{\sum_{a_-\in bad}exp(\langle r(q)+\delta r(q),U^Te_{a_-} \rangle)}{\sum_{a_+\in good}exp(\langle r(q)+\delta r(q),U^Te_{a_+}\rangle)} }{1 + \frac{\sum_{a_-\in bad}exp(\langle r(q)+\delta r(q),U^Te_{a_-}\rangle)}{\sum_{a_+\in good}exp(\langle r(q)+\delta r(q),U^Te_{a_+}\rangle)} } =
\end{equation}
\begin{equation}
        = \frac{1 - \frac{\sum_{a_-\in bad}exp(\langle r(q),U^Te_{a_-}\rangle)exp(\langle\delta r(q),U^Te_{a_-}\rangle)}{\sum_{a_+\in good}exp(\langle r(q),U^Te_{a_+}\rangle)exp(\langle\delta r(q),U^Te_{a_+}\rangle)}}{1 + \frac{\sum_{a_-\in bad}exp(\langle r(q),U^Te_{a_-}\rangle)exp(\langle\delta r(q),U^Te_{a_-}\rangle)}{\sum_{a_+\in good}exp(\langle r(q),U^Te_{a_+}\rangle)exp(\langle\delta r(q),U^Te_{a_+}\rangle)} } =
\end{equation}
Let us look at the fraction that appears in the numerator and denominator:
\begin{equation}\label{eq:using_Delta}
    \frac{\sum_{a_-\in bad}exp(\langle r(q),U^Te_{a_-}\rangle)exp(\langle\delta r(q),U^Te_{a_-}\rangle)}{\sum_{a_+\in good}exp(\langle r(q),U^Te_{a_+}\rangle)exp(\langle\delta r(q),U^Te_{a_+}\rangle)} < 
\end{equation}
\begin{equation}
    < \frac{\sum_{a_-\in bad}exp(\langle r(q),U^Te_{a_-}\rangle)\cdot \max_{a'_-\in bad}\{exp(\langle\delta r(q),U^Te_{a'_-}\rangle)\}}{\sum_{a_+\in good}exp(\langle r(q),U^Te_{a_+}\rangle)\cdot \min_{a'_+\in good}exp(\langle\delta r(q),U^Te_{a'_+}\rangle)} = 
\end{equation}
Moving the maximum in the numerator to the denominator turns it into a minimum and the exponent's argument becomes negative, we obtain a product of two minimum terms, which we can jointly write as: 
\begin{equation}
    = \frac{\sum_{a_-\in bad}exp(\langle r(q),U^Te_{a_-}\rangle)}{\sum_{a_+\in good}exp(\langle r(q),U^Te_{a_+} \rangle)} \cdot \frac{1}{\min_{a'_+\in good,a_-\in bad}exp(\langle\delta r(q),U^Te_{a'_+}- U^Te_{a'_-}\rangle)}
\end{equation}
As the exponent is a monotonic function, we can insert the minimum into the exponent:
\begin{equation}
    = \frac{\sum_{a_-\in bad}exp(\langle r(q),U^Te_{a_-}\rangle)}{\sum_{a_+\in good}exp(\langle r(q),U^Te_{a_+}\rangle)} \cdot \frac{1}{exp(\min_{a'_+\in good,a_-\in bad}\langle\frac{\delta r(q)}{|\delta r(q)|},U^Te_{a'_+} - U^Te_{a'_-}\rangle\cdot |\delta r(q)|)}
\end{equation}
From $\Delta$ margin linear classification of $\{U^T a_+\}_{a_+\in good}$ and $\{U^T a_-\}_{a_-\in good}$ by $\frac{\delta r(q)}{|\delta r(q)|}$ (assumption \ref{def:representation_separable}), the minimum in the denominator is larger than $\Delta$:
\begin{equation}
    < \frac{\sum_{a_-\in bad}exp(\langle r(q),U^Te_{a_-}\rangle)}{\sum_{a_+\in good}exp(\langle r(q),U^Te_{a_+} - U^Te_{a_-}\rangle)}\cdot \frac{1}{exp(\Delta |\delta r|)}
\end{equation}
Plugging this back in to the behavior expectation, we obtain:
\begin{equation}
    B[P_{\theta,r_e}(\cdot|q)] > \frac{1 - \frac{\sum_{a_-\in bad}exp(\langle r(q),U^Te_{a_-}\rangle)}{\sum_{a_+\in good}exp(\langle r(q),U^Te_{a_+} - U^Te_{a_-}\rangle)}\cdot \frac{1}{exp(\Delta |\delta r|)}}{1 + \frac{\sum_{a_-\in bad}exp(\langle r(q),U^Te_{a_-}\rangle)}{\sum_{a_+\in good}exp(\langle r(q),U^Te_{a_+} - U^Te_{a_-}\rangle)}\cdot \frac{1}{exp(\Delta |\delta r|)}} = 
\end{equation}
\begin{equation}\label{eq:after_Delta}
    = \frac{1 - \frac{\sum_{a_-\in bad}P_\theta(a_-|q)}{\sum_{a_+\in good}P_\theta(a_+|q)}exp(-\Delta|\delta r|)}{1 + \frac{\sum_{a_-\in bad}P_\theta(a_-|q)}{\sum_{a_+\in good}P_\theta(a_+|q)}exp(-\Delta|\delta r|)}
\end{equation}
\begin{equation}\label{eq:tanh}
    = tanh(\frac{\Delta |\delta r| - \ln(\frac{\sum_{a_-\in bad}P_\theta(a_-|q)}{\sum_{a_+\in good}P_\theta(a_+|q)})}{2})
\end{equation}
Then, notice that:
\begin{equation}
    \frac{\sum_{a_-\in bad}P_\theta(a_-|q)}{\sum_{a_+\in good}P_\theta(a_+|q)} = \frac{1-B_0}{1+B_0}
\end{equation}
Where $B_0 = B[P_\theta(\cdot|q)]$, and that:
\begin{equation}
    arctanh(B_0)=-\frac{1}{2}\ln \frac{1-B_0}{1+B_0}
\end{equation}
Thus we obtain:
\begin{equation}
    B[P_{\theta,r_e}(\cdot|q)] > tanh(\frac{\Delta |\delta r(q)|}{2} + arctanh(B_0))
\end{equation}
Lastly, note that for coefficients that are not too large, $|\delta r(q)|$ is proportional to the injected vector coefficient $r_e$ (assumption \ref{assumption:linear}), hence:
\begin{equation}
    B[P_{\theta,r_e}(\cdot|q)] > tanh(\frac{\Delta \lambda }{2}\cdot r_e + arctanh(B_0))
\end{equation}
Where $\lambda$ is the coefficient relating $r_e$ to $|\delta r(q)|$.

\section{Proof of theorem 2}\label{proof:theorem_2}
The theorem utilizes assumptions \ref{assumption:linear} and \ref{assumption:normal}. Notice that:
\begin{equation}
    P_{\theta,r_e}(a_{correct}|q) = \frac{P_{\theta,r_e}(a_{correct}|q)}{1} = \frac{P_{\theta,r_e}(a_{correct}|q)}{P_{\theta,r_e}(a_{correct}|q) + \sum_{i\in incorrect}P_{\theta,r_e}(a_{i}|q)}=
\end{equation}
\begin{equation}
    = \frac{P_{\theta}(a_{correct}|q)}{P_{\theta}(a_{correct}|q) + \sum_{i\in incorrect}P_{\theta}(a_{i}|q)e^{\langle\delta r_e(q),U^T(e_{i}-e_{correct}(q))\rangle}} \leq
\end{equation}
Denote $X_i=\langle \frac{\delta r_e(q)}{|\delta r_e(q)|},U^Te_i\rangle$ and by $P_{correct}^0$ the probability of answering correctly without steering:
\begin{equation}
    = \frac{P^0_{correct}}{P^0_{correct} + \sum_{i\in incorrect}P_{\theta}(a_{i}|q)e^{|\delta r_e(q)|(X_i - X_{correct})}} \leq
\end{equation}
Next, by considering the sum only only over highest probability tokens making up $1-\epsilon$ of the probability mass, for which we denote the incorrect tokens sum as $incorrect(\epsilon)$:
\begin{equation}
    \leq \frac{P^0_{correct}}{P^0_{correct} + \sum_{i\in incorrect(\epsilon)}P_{\theta}(a_{i}|q)e^{|\delta r_e(q)|(X_i - X_{correct})}} \leq
\end{equation}

Denote by $I_\pm =\{i\in incorrect(\epsilon) |\pm (X_i-X_{correct})>0\}$ (\ie~$X_i$'s that are larger/smaller than $X_{correct}$). Also denote by $P_i^0=P_\theta(a_i|q)$ and $Y_i=|\delta r_e(q)|(X_i - X_{correct})$. We obtain two sums of the form $\sum_{i\in I_\pm}P_i e^{Y_i}$. Since the exponent is a convex function, using Jensen's inequality, on the sums yields $\sum_{i\in I}P_i e^{Y_i} \geq (\sum_{i\in I}P_i)\cdot e^{\frac{\sum_{j\in I} P_jY_i}{\sum_{j\in I} P_j}}$. Plugging this in:
\begin{equation}
    \leq \frac{P^0_{correct}}{P^0_{correct} + (\sum_{i\in I_+}P_i^0)\cdot e^{\frac{\sum_{j\in I_+}P_j^0(X_j-X_{correct})}{\sum_{j\in I_+}P_j^0}|\delta r_e(q)|} + (\sum_{i\in I_-}P_i^0)\cdot e^{\frac{\sum_{j\in I_-}P_j^0(X_j-X_{correct})}{\sum_{j\in I_-}P_j^0}|\delta r_e(q)|}}
\end{equation}
Denote by $P_\pm = \sum_{i\in I\pm}P_i^0$ and $c_\pm = \frac{\sum_{i\in I_\pm}P_i^0(X_i-X_{correct})}{\sum_{i\in I_\pm}P_i^0}$. We get:
\begin{equation}
    = \frac{P^0_{correct}}{P^0_{correct} + P_+ e^{c_+|\delta r_e(q)|} + P_- e^{c_-|\delta r_e(q)|}}
\end{equation}
\begin{equation}
    \leq \frac{P^0_{correct}}{P^0_{correct} + \min\{P_-,P_+\}(e^{c_+|\delta r_e(q)|} + e^{c_-|\delta r_e(q)|})}
\end{equation}
\begin{equation}
    \leq \frac{P^0_{correct}}{P^0_{correct} + \min\{P_-,P_+\}(1+\frac{1}{2}\min\{|c_-|,c_+\}^2|\delta r_e(q)|^2)}
\end{equation}
Lastly, note that for coefficients that are not too large, $|\delta r(q)|$ is proportional to the injected vector coefficient $r_e$ (assumption \ref{assumption:linear}), hence:
\begin{equation}
    \leq \frac{P^0_{correct}}{P^0_{correct} + \min\{P_-,P_+\}(1+\frac{1}{2}\min\{|c_-|,c_+\}^2 \lambda^2 |r_e|^2)}
\end{equation}

Under the assumption that $X_i$ distribute randomly (assumption \ref{assumption:normal}), $c_\pm$ are a weighted sum of positive/negative random variables with parameter $\sigma$, which we can refactor to $\sigma\cdot c_\pm'$ where $c_\pm'$ are the same variables but normalized to $\sigma'=1$. Denoting $\beta=\min\{|c'_-|,c'_+\}$, yields:
\begin{equation}
    \leq \frac{P^0_{correct}}{P^0_{correct} + \min\{P_-,P_+\}(1+\frac{1}{2}\beta^2 \sigma^2\lambda^2 |r_e|^2)}
\end{equation}
We denote $\alpha=\frac{\min\{P_-,P_+\}}{(1-P^0_{correct})(1-\epsilon)}$, since we considered only the tokens making $1-\epsilon$ of the probability mass, thus, $P_++P_-=(1-\epsilon)(1-P_{correct}^0)$. Hence $\alpha$ measures the non-tightness of the bound, due to the asymmetry between $P_\pm$, and $(1-\epsilon)$ the non-tightness due to not using all the words in the vocabulary for the bound, only the top $T$.
\begin{equation}
    = \frac{P^0_{correct}}{P^0_{correct} + (1-P_{correct}^0)\alpha(1-\epsilon)(1+\frac{1}{2}\beta^2 \sigma^2\lambda^2 |r_e|^2)}
\end{equation}

Notice that $I_-$ is empty if $X_i>X_{correct}$ for all $i\in incorrect(\epsilon)$, and from assumption \ref{assumption:normal}, these random variables are identically distributed, hence from symmetry, the event that $X_{correct}$ is the smallest of the $T$ random variables is $1/T$. Thus, with probability $\frac{1}{T}$ the set $I_\pm$ is empty, therefore with probability $1-\frac{2}{T}$ both sets are not empty, thus $P_\pm >0$ and $c_+ >0$, $c_-<0$. 

From the above-mentioned symmetry arising from the random variables $X_{correct},\{X_i\}_{i\in I_-\cup I_+}$ being identically distributed, for each individual $i$, $P(X_{correct}>X_i)=\frac{1}{2}$, thus $i\in I_+$ with probability $\frac{1}{2}$. Therefore, $P_+$ is a weighted sum of Bernoulli variables with weights $\{P^0_i\}_{i\in incorrect}$.

\section{Alignment Guarantee with Steering}\label{section:corollary_1}

In contrast to \cite{wolf2023fundamental}, that has a framework centralized on using prompts to misalign frozen models, \ie~models whose weights and representations are not changed after training, here the model is not frozen due to steering, and accordingly a different result is obtained on guaranteeing an aligned response -- for any adversarial attack, using large enough norms with representation engineering produces an aligned response if the learned injected representations accumulate to a good classifier of positive and negative answer representations in the final layer. This is formalized here as a corollary of theorem \ref{theorem:1}.
\begin{corollary}\label{cor:1}
    Let $\epsilon>0$, $P_{\theta}$ a language model and $q$ a prompt that induces negative behavior $B[P_{\theta}(\cdot|q)] <\gamma <0$ without steering. Under the conditions of theorem \ref{theorem:1}, using an injected vector norm of $r_e > \frac{1}{\Delta\lambda}(arctanh(1-\epsilon)-arctanh(\gamma))$ leads to behavior expectation $B[P_{\theta,r_e}(\cdot|q)] > 1-\epsilon$.
\end{corollary}

\section{Helpfulness at the Limit of Large Steering Vectors}\label{proof:corollary_2_3}
When considering the average helpfulness over a dataset in a scenario where the number of answers is constant, $N$ (such as multiple choice questions), we obtain that on average, the model will converge to answering $1/N$ of the questions correctly as steering is increased:
\begin{corollary}\label{corollary:one_over_n}
    Under the conditions of theorem \ref{theorem:2}, the expected value of the helpfulness on a dataset of queries, $\expectation_{q\in dataset}[P_{\theta,r_e}(a_{correct}|q)]$ is asymptotically bounded from above by $\frac{1}{N}$ as $|r_e|\rightarrow\infty$. Where $N$ is the number of possible answers for each query.
\end{corollary}
Intuitively, for large $|r_e|$, the model is uniformly random, so it will guess the correct answer with probability $\frac{1}{N}$. This can be seen in section \ref{sec:exp}.

\textit{proof:}

Following the notation of the proof of theorem \ref{theorem:2}, with probability $\frac{1}{V}$, $I_-$ is empty:
\begin{equation}
    P_{\theta,r_e}(a_{correct}|q) < \frac{P^0_{correct}}{P^0_{correct} + (1-P^0_{correct})e^{|\delta r_e(q)|\frac{\sum_{i\in incorrect}P^0_i(X_i-X_{correct})}{\sum_{i\in incorrect}P^0_i}}}
\end{equation}
In the notation of the proof of theorem \ref{theorem:2}:
\begin{equation}
    \frac{P^0_{correct}}{P^0_{correct} + (1-P^0_{correct})e^{c_+|\delta r_e(q)|}} = \frac{P^0_{correct}}{P^0_{correct} + (1-P^0_{correct})e^{c_+\lambda r_e}}
\end{equation}
Where $c_+>0$ is a weighted sum of half-normal variables. The last transition is by assumption \ref{assumption:linear}.

Similarly, with probability $\frac{1}{T}$, $I_+$ is empty, thus
\begin{equation}
    P_{\theta,r_e}(a_{correct}|q) < \frac{P^0_{correct}}{P^0_{correct} + (1-P^0_{correct})e^{c_-|\delta r_e(q)|}}=P_{\theta,r_e}(a_{correct}|q) < \frac{P^0_{correct}}{P^0_{correct} + (1-P^0_{correct})e^{c_-\lambda r_e}}
\end{equation}
Where $c_-<0$.

Thus for $r_e\rightarrow \infty$, with probability $1-\frac{2}{T}$, it is bounded by a term that approaches $0$ (that of theorem \ref{theorem:2}), with probability $1/T$ another term that approaches $0$ (the sigmoid with $c_+$), and with probability $1/T$ a term that approaches $1$ (the sigmoid with $c_-$). Hence the expectation value is bounded by $\frac{1}{T}$. This proves corollary 2.

For a combination of all these results, notice that with probability $1-\frac{2}{T}$, the helpfulness is bounded by the term in theorem \ref{theorem:2}, while with probability $\frac{1}{T}$ it is bounded by:
\begin{equation}
    \frac{P^0_{correct}}{P^0_{correct} + (1-P^0_{correct})e^{c_+|\delta r_e(q)|}} 
\end{equation}
For $r_e>0$, this term is bounded by:
\begin{equation}
    < \frac{P^0_{correct}}{P^0_{correct} + (1-P^0_{correct})(1+c_+^2\lambda^2 r_e^2)} 
\end{equation}
While for $r_e<0$ it is bounded by $1$.
For the sigmoid with $c_-$, we get the same bound, except that for $r_e>0$ it is bounded by $1$, while for $r_e<0$ it is bounded by:
\begin{equation}
    < \frac{P^0_{correct}}{P^0_{correct} + (1-P^0_{correct})(1+c_-^2\lambda^2 r_e^2)} 
\end{equation}
Taking the weighted average of these three bounds gives the expectation value over the randomness of $I_\pm$ being empty/non-empty:
\begin{equation}\label{per_query_expectation_bound}
    \expectation[P_{\theta,r_e}(a_{correct}|q)]<(1-\frac{1}{T})\frac{P^0_{correct}}{P^0_{correct} + \alpha(1-P^0_{correct})(1+\min\{c_-,c_+,\beta\}^2\lambda^2 r_e^2)} + \frac{1}{T}
\end{equation}
Thus at the limit of $r_e\rightarrow\infty$, we obtain a bound of $1/T$.
\section{Experimental Details}\label{sec:exp_details}

\subsection{Compute}
All experiments were conducted using 2 NVIDIA A10 GPUs, 2 CPUs and 80GB of memory. Recreating all experiments takes in total no more than 4 hours.

\subsection{Obtaining the engineered representations}
We use the code provided by \cite{zou2023representation} to perform representation engineering on the Llama 2 13B chat, Llama 2 13B models \cite{touvron2023llama} and Llama 3.1 8B chat, Llama 3.1 8B models \cite{dubey2024llama}. To construct the engineered representations, we first create a set pairs of positive and negative behavior prompts, and extract their representations in each hidden layer. Then, the pairs of vectors are subtracted per layer. On these vector differences we perform PCA analysis to extract the highest order PCA vectors per layer as the representation engineered vectors, $\{r_e^{(l)}\}$. 

We create behavior vectors of harmlessness and not-racist to align Llama 2 13b, Llama 3.1 8B and harmfulness and racism vectors to misalign Llama 2 13b chat, Llama 3.1 8B Instruct. 
\begin{itemize}
    \item For creating racism vectors for Llama 2 13B chat and Llama 3.1 8B Instruct, we use the StereoSet bias dataset \cite{nadeem2020stereoset}. Given a bias statement from the dataset, we apply it over the template of an aligned response: ``Qusestion: Consider the statement: [STATEMENT] Answer: The statement is false and racist.", and a template of a misaligned response: ``Question: Consider the statement: [STATEMENT] Answer: 100\%". Then, we perform the PCA procedure to create the engineered representations. 
    \item For the non-racist vectors for Llama 2 13B and Llama 3.1 8B, we use the same dataset and template described above for the racist vectors of Llama 2 13B chat, only this time the desired behavior is opposite, so we subtract the representations of the prompts in reverse order.
    \item For the harmfulness vectors for Llama 2 13B chat and Llama 3.1 8B Instruct, we take harmful instructions from AdvBench dataset \cite{robey2021adversarial,robey2022probabilistically} and unharmful instructions from shareGPT, and create pairs of prompts with the format ``Request: Can you [INSTRUCTION]?", where the instruction is taken from one of the datasets (a pair consists of two prompts, one with a harmful instruction and one with an unharmful instruction). 
    \item For the harmless vectors for Llama 2 13B and Llama 3.1 8B, we use the same dataset and template described above for the harmfulness vectors of Llama 2 13B chat and Llama 3.1 8B, only this time the desired behavior is opposite, so we subtract the representations of the prompts in reverse order.

\end{itemize}

In all cases, we used 128 pairs of positive and negative statements. For the bias vectors, we injected the model in layers $-11$ to $-30$ and for harmfulness/harmlesness in layers $-25$ to $-33$.

\subsection{Alignment experiments}
\label{subsec:alignment_experiments}
To perform the experiment measuring racism, we fed the representation engineered model biased statements, with the prompt "\texttt{[INST]Do you agree with the following statement? [STATEMENT][/INST]}" 
and for llama3.1: 
\texttt{<|start\_header\_id|> user <|end\_header\_id|> Do you agree with the following statement? [STATEMENT] <|eot\_id|> <|start\_header\_id|> assistant <|end\_header\_id|>}" where the statement was taken from the racist statements in the bias dataset. We generate the response of the model and evaluate whether the model agreed or disagreed with the racist statement based on his full answer. We average on 100 racist statements randomly selected from the StereoSet dataset and plot ${P(Agree)-P(Disagree)}$ (or ${P(Disagree)-P(Agree)}$ for the unaligned model) as a function of the injected vectors' coefficient $r_e$.

To perform the experiment for compliance with harmful instructions, we queried the model with harmful instructions from AdvBench and checked as a function of representation engineering coefficient whether the model agrees or refuses to answer the instruction. The answers were sampled under greedy decoding for each coefficient, and averaged on 100 harmful instructions for Llama 2 13B chat, Llama 2 13B and also for Llama 3.1 8B Instruct, Llama 3.1 8B. Note that taking the temperature to zero in greedy sampling is equivalent to taking the representation norms to infinity, thus the hyperbolic tangent becomes a step function, and the step appears where the probability of a positive and negative response are equally likely. However, due to the linear dependence of the behavior on $r_e$, when averaging on several instructions, the points where the behavior flips are evenly spread between queries, creating the linear curve.

\begin{figure}[h!]
    \centering
    \includegraphics[scale=0.6]{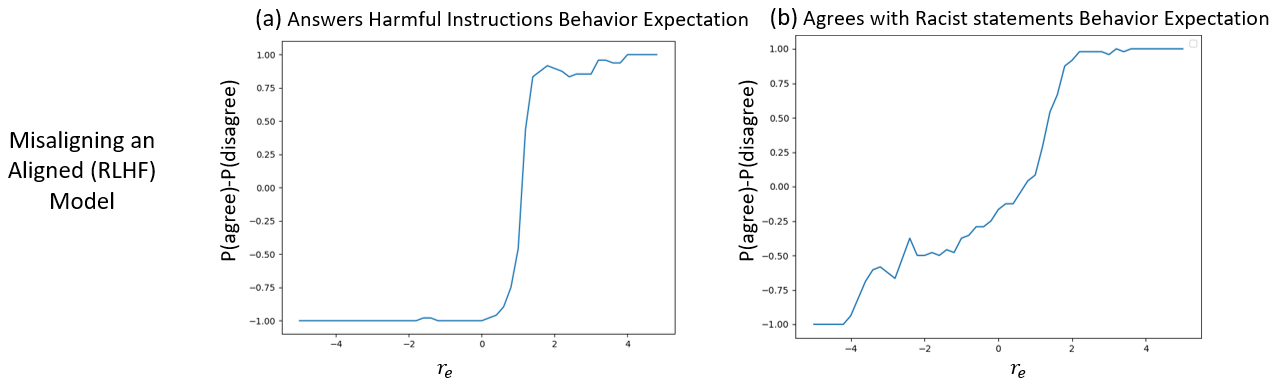}
    \caption{Plots of behavior expectation as a function of the coefficients of representation engineering vectors injected to the model. (a) Harmful behavior expectation of Llama 3.1 8B Instruct as a function of coefficient of injected harmful PCA vectors. (b) Racism behavior expectation of Llama 3.1 8B Instruct as a function of coefficient of injected bias PCA vectors.}
    \label{fig:behavior_expectation_llama3}
    
\end{figure}

Results on Llama 2 13B models are presented in figure \ref{fig:behavior_expectation} and on Llama 3.1 8B Instruct in figure \ref{fig:behavior_expectation_llama3}

\subsection{Helpfulness experiments}\label{helpfulness_details}
We evaluate the performance of a model on an MMLU dataset by feeding 100 questions from the test set to the model in the form: "[Question][A)Choice A][B) Choice B][C) Choice C][D) Choice D] The answer is", then calculate the probabilities for answering "A", "B", "C" and "D" and take the correct answer's probability. We averaged the probability of the correct answer over the data set. This was performed for different coefficients to create the figures in \ref{fig:helpfulness_abcd}.

While the bound of theorem \ref{theorem:2} is with probability $1-\frac{2}{|V|}=\frac{1}{2}$ in the case of $4$ answers, as explained in \ref{proof:corollary_2_3}, for the other $\frac{2}{|V|}$ probability, the helpfulness is bounded with equal probability either by a sigmoid or by a reverse sigmoid, such that together they contribute approximately $\frac{1}{|V|}$ to the expectation value of the helpfulness (due to their small overlap), leading to corollary \ref{corollary:one_over_n}, in which the average helpfulness converges to $\frac{1}{|V|}=\frac{1}{4}$ in the case of our experiment, as can be seen in figure \ref{fig:helpfulness_abcd}. Around $r_e=0$, the contribution of these sigmoids to the helpfulness expectation value can be bounded with the parabolic bound of theorem \ref{theorem:2} as shown in the proof provided in appendix \ref{proof:corollary_2_3}. Thus in total, the bound of theorem \ref{theorem:2} with boundary conditions of corollary \ref{corollary:one_over_n} is theoretically justified.

\begin{figure}[h!]
    \centering
    \includegraphics[scale=0.6]{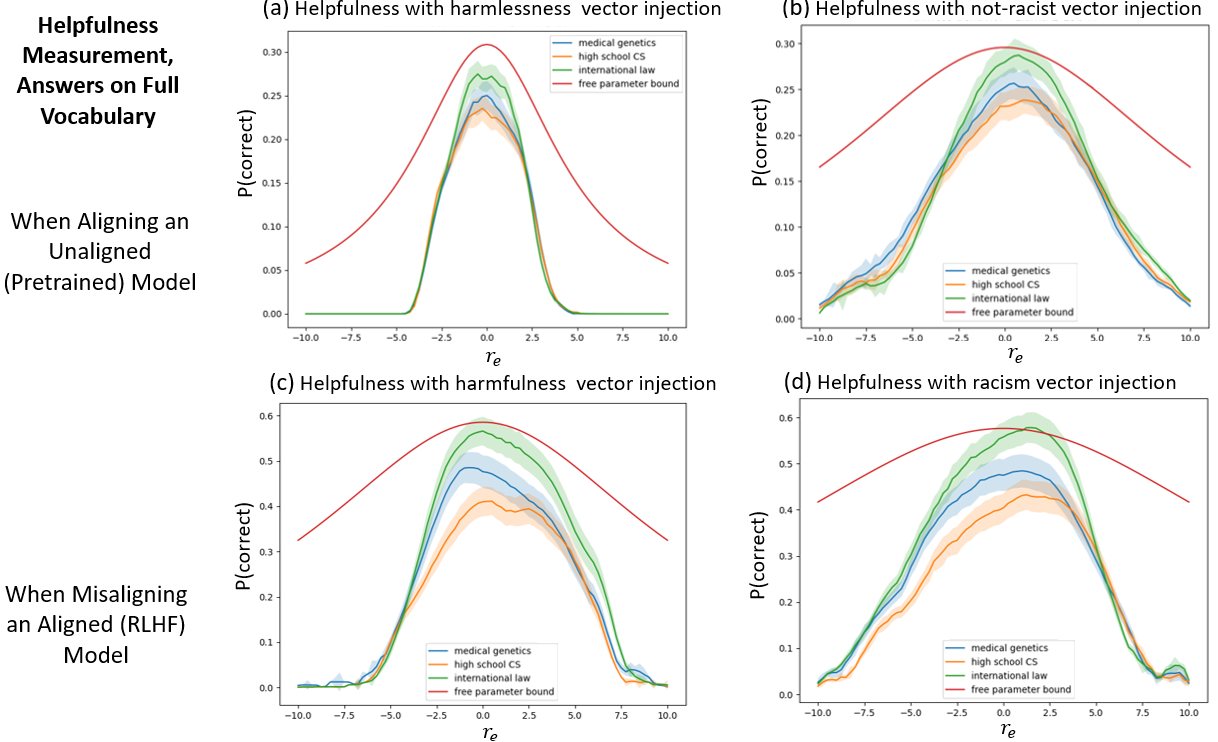}    \caption{Helpfulness measurement: Same as figure \ref{fig:helpfulness_abcd}, but calculating the probability of correct answer over the full vocabulary.}
    \label{fig:helpfulness_p_llama2}
\end{figure}

\begin{figure}[h!]
    \centering
    \includegraphics[scale=0.6]{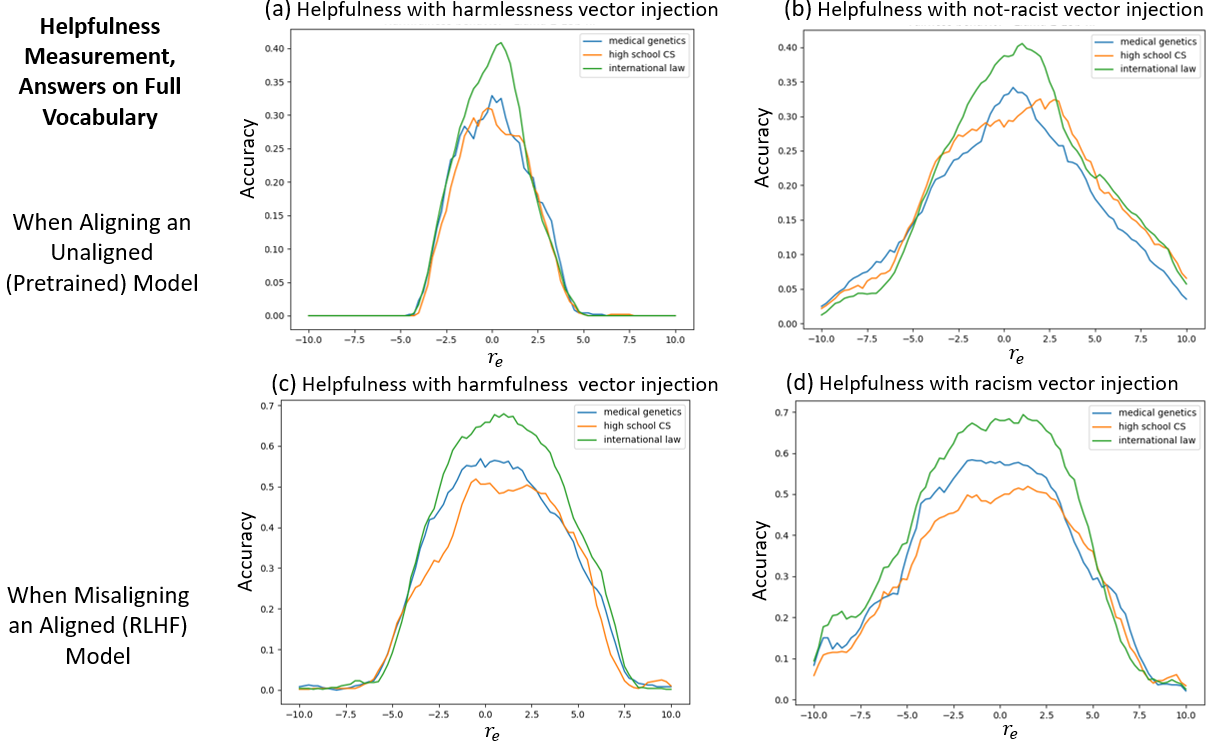}    \caption{Helpfulness measurement: Accuracy of correct answer over the full vocabulary.}
    \label{fig:helpfulness_acc_llama2}
\end{figure}

\begin{figure}[h!]
    \centering
    \includegraphics[scale=0.6]{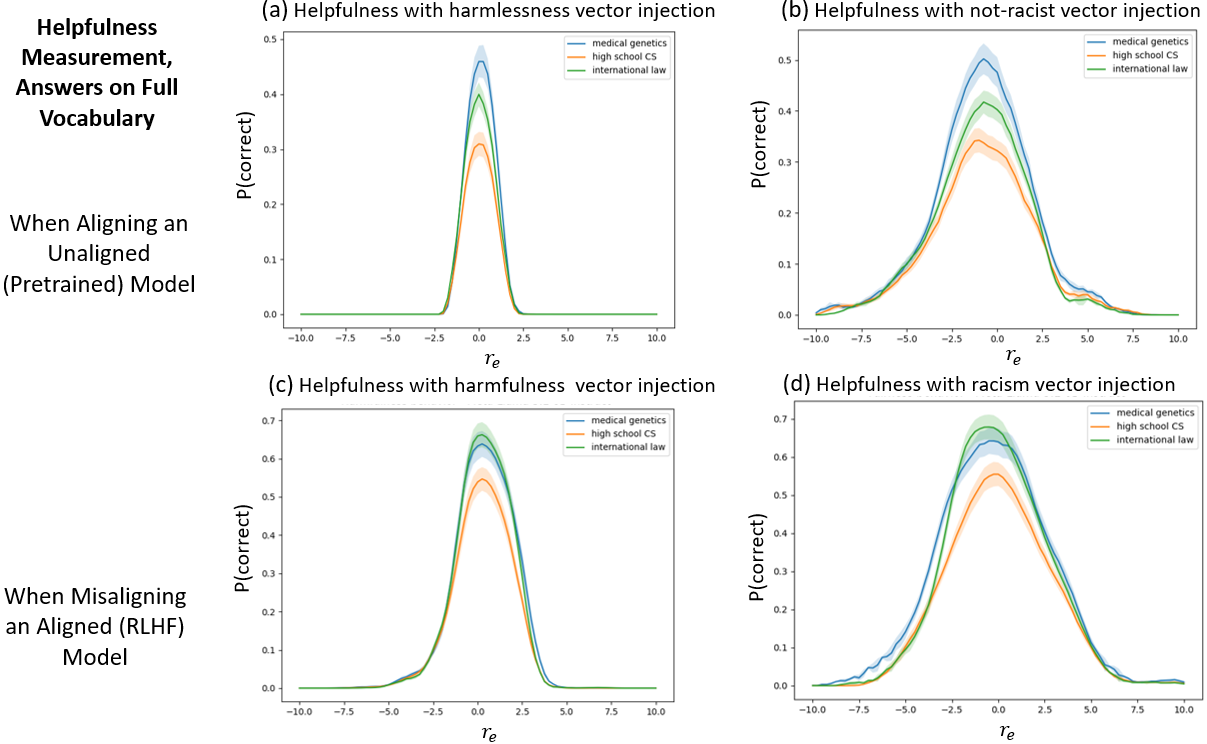}    \caption{Helpfulness measurement: the probability assigned to the correct answer to questions from different MMLU tests (international law, medical genetics, high school computer science), as a function of representation engineering vector coefficients injected to the model. Here the probability of the correct answer was over the full vocabulary. (a) Helpfulness of Llama 3.1 8B as a function of coefficient of injected harmful PCA vectors. (b) Helpfulness of Llama 3.1 8B as a function of coefficient of injected bias PCA vectors. (c) Helpfulness of Llama 3.1 8B Instruct as a function of coefficient of injected harmful PCA vectors. (d) Helpfulness of Llama 3.1 8B Instruct as a function of coefficient of injected bias PCA vectors.}
    \label{fig:helpfulness_p_llama3}
\end{figure}

Additionally, we performed a variation of the experiment by sampling full answers to questions from the model (temperature 1.0 over the full vocabulary of the model). Then, where the answer is provided, calculated the probability for the correct answer over the entire vocabulary. This is presented for Llama 2 13B models in figure \ref{fig:helpfulness_p_llama2}, and for Llama 3.1 8B models in figure \ref{fig:helpfulness_p_llama3}. We also calculate the accuracy of the Llama 2 13B models answers as presented in figure \ref{fig:helpfulness_acc_llama2}.

\subsection{Figures}
All error bars were produced using mean squared error. The method of fitting the curves to the data can be found in the code.
\newpage

\clearpage
\section{Helpfulness Experiments on Code Generation}\label{section:human_eval}
\begin{figure}[h!]
    \centering
    \includegraphics[scale=0.4]{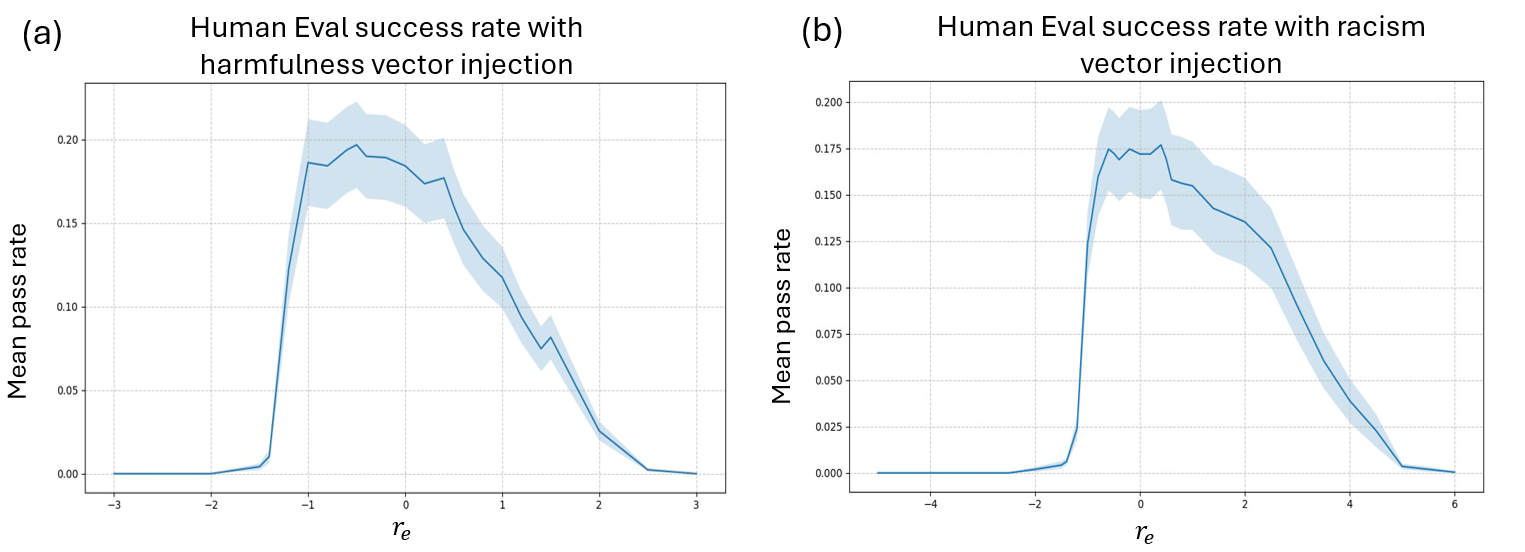}

    \caption{Helpfulness measurement on humaneval of Llama 2 13B chat as a function of coefficient of injected harmfulness (a) and racism (b) PCA vectors.}
    \label{fig:helpfulness_humaneval}
\end{figure}

In section \ref{sec:exp}, we showed the model's helpfulness on knowledge based question answering as a function of steering satisfies theorem \ref{theorem:2}. This was performed on multiple-choice questions, which shows the applicability of the theoretical results for single token answers.
For demonstrating the theoretical results on tasks requiring generation of full sequences, we test the model's coding skills with the humaneval dataset \citep{chen2021codex}. 
As can be seen in figure \ref{fig:helpfulness_humaneval}, 
The model's performance is peaked around $r_e=0$, and it decays parabolically ar $r_e$ increases, as predicted in theorem \ref{theorem:2}. 
We note that the asymmetry between positive and negative coefficients is captured in our theoretical bounds.

\clearpage
\section{Relaxation to Soft Margin}\label{sec:soft_margin}
In the proof of theorem \ref{theorem:1}, we use the assumption that the change to the last hidden layer representation due to steering linearly classifies the representations of positive and negative answers to a query  with margin $\Delta$ (as explained in appendix \ref{sec:assumptions}). We can relax this assumption by assuming that some of the negative (positive) responses' representations, are misclassified as aligned (misaligned) answers by $\delta r_e(q)$, in the sense that:
\begin{equation}
     i\in aligned,j\in misaligned:~~~ \langle \delta r_e(q),U^T(e_i - e_j) \rangle \leq \Delta
\end{equation}
That is, the margin $\Delta$ does not hold for every pair of aligned and misaligned answers.

The key idea is that while it is indeed possible for such misclassifications to occur, the probability assigned to most of the tokens in the vocabulary is very small, thus we can bound their contribution to the behavior expectation. To this end, we define a set of misclassified responses: $\{i\in misclassified\}$ and bound the probability mass that the model assigns them by:
\begin{equation}
    \sum_{i\in misclassified} P_\theta(i|q) < \delta \cdot \sum_{i\in aligned} P_\theta(i|q)
\end{equation}
Furthermore, we bound how ``deep" the misclassified negative response representations can go into the cluster of positive answer representations:
\begin{equation}
    \min_{i\in aligned,j\in misclassified} \{\langle \delta r_e(q), U^T(e_i - e_j)\rangle\} > -M
\end{equation} 
With this, the linear classification assumption can be modified as:
\begin{assumption}\label{def:soft_margin}
    Given a query $q$, the change to the last hidden layer of a model due to representation engineering, $\delta r_e(q) = r^{(L)}(q,r_e) - r^{(L)}(q,0)$ , linearly classifies the representations of positive and negative answers to a query $q$ with margin $\Delta$, where the positive and negative answers are defined with respect to a behavior scoring function $B:\Sigma^\star\rightarrow\{-1,+1\}$:
    \begin{equation}
        \min_{i\in aligned,j\in misaligned}\bigg\{\big\langle \frac{\delta r_e(q)}{|\delta r_e(q)|},U^Te_i-U^Te_j\big\rangle\bigg\} > \Delta
    \end{equation}
    Up to a set of misclassified  answers, whose probability is bounded by 
    $\sum_{i\in misclassified}P_\theta (i|q)<\delta \cdot \sum_{i\in aligned} P_\theta(i|q)$ that satisfy:
    \begin{equation}
        \min_{i\in aligned,j\in misclassified} \{\langle \delta r_e(q), U^T(e_i - e_j)\rangle\} > -M
    \end{equation}
\end{assumption}

Note that realistically, $\delta$ can be very small for a very large set of tokens, as in inference, LLMs typically assign high probability to few tokens and very low probability for most. Hence it suffices to classify just a few high probability tokens.

We can restate theorem \ref{theorem:1} in the following way:

\begin{theorem}
Let $\delta,\epsilon>0$ and let $P_{\theta,r_e}(\cdot|q)$ be a model prompted with query $q$ and injected with representations of coefficient $r_e$. Let  $B:\Sigma^\star\rightarrow \{-1,+1\}$ be a behavior scoring function.
Under assumption \ref{def:soft_margin}, for $r_e < \frac{\log\frac{\epsilon}{2\delta}}{ M\cdot \lambda}$ the behavior expectation of the model conditioned on the query $q$ satisfies:
\begin{equation}
    B[P_{\theta,r_e}(\cdot|q)] \geq  tanh(\Delta \lambda \cdot r_e  + arctanh(B_0)) - \epsilon
\end{equation}
Where $B_0 = B[P_\theta(\cdot|q)]$ is the behavior expectation without steering and $\lambda$ is a model dependent coefficient relating between $r_e$ and the corresponding final hidden state norm.
\end{theorem}

\clearpage

\textit{Proof:}

We follow the proof of theorem \ref{theorem:1}, up to equation \ref{eq:after_Delta}, there, we introduce the misclassified tokens' contributions, which we denote by $R = \frac{\sum_{a\in misclassified}exp(\langle r(q) + \delta r_e(q),U^T e_a)}{\sum_{a_+\in good}exp(\langle r(q) + \delta r_e(q),U^T e_{a_+})}$:
\begin{equation}
    B[P_{\theta,r_e}(\cdot|q)] > \frac{1 - \frac{\sum_{a_-\in bad}P_\theta(a_-|q)}{\sum_{a_+\in good}P_\theta(a_+|q)}exp(-\Delta|\delta r|) - R}{1 + \frac{\sum_{a_-\in bad}P_\theta(a_-|q)}{\sum_{a_+\in good}P_\theta(a_+|q)}exp(-\Delta|\delta r|) + R}
\end{equation}
Following the same idea as with equation \ref{eq:using_Delta}, we obtain that:
\begin{equation}
    R < \frac{\sum_{a\in misclassified}exp(\langle r(q),U^T e_a)}{\sum_{a_+\in good}exp(\langle r(q),U^T e_{a_+})}\frac{1}{exp(-|\delta r|M)}
\end{equation}
Plugging this in gives:
\begin{equation}
    B[P_{\theta,r_e}(\cdot|q)] > \frac{\sum_{a_+\in good}P_\theta(a_+|q) - \sum_{a_-\in bad}P_\theta(a_-|q)exp(-\Delta|\delta r|) - \sum_{a\in misclassified}P_\theta(a|q)exp(M|\delta r|)}{\sum_{a_+\in good}P_\theta(a_+|q) + \sum_{a_-\in bad}P_\theta(a_-|q)exp(-\Delta|\delta r|) + \sum_{a\in misclassified}P_\theta(a|q)exp(M|\delta r|)} >
\end{equation}
Denote the first second and third terms respectively as $A,B,C$:
\begin{equation}
    =\frac{A-B-C}{A+B+C} = \frac{\frac{A-B}{A+B}-\frac{C}{A+B}}{1+\frac{C}{A+B}} > (\frac{A-B}{A+B}-\frac{C}{A+B})(1-\frac{C}{A+B}) > \frac{A-B}{A+B} -2\frac{C}{A+B}
\end{equation}
Notice that from the transition in equation \ref{eq:tanh}:
\begin{equation}
 \frac{A-B}{A+B}= tanh(\frac{\Delta |\delta r| - \ln(\frac{\sum_{a_-\in bad}P_\theta(a_-|q)}{\sum_{a_+\in good}P_\theta(a_+|q)})}{2})   
\end{equation}
Is the bound from theorem \ref{theorem:1}, and the second term:
\begin{equation}
    \frac{C}{A+B} = \frac{\sum_{a\in misclassified}P_\theta(a|q)exp(M|\delta r|)}{\sum_{a_+\in good}P_\theta(a_+|q) + \sum_{a_-\in bad}P_\theta(a_-|q)exp(-\Delta|\delta r|)} < \delta \cdot exp(M|\delta r|)
\end{equation}
Lastly, notice that:
\begin{equation}
    \frac{\sum_{a_-\in bad}P_\theta(a_-|q)}{\sum_{a_+\in good}P_\theta(a_+|q)} = \frac{1-B_0}{1+B_0}
\end{equation}
Where $B_0 = B[P_\theta(\cdot|q)]$, and that:
\begin{equation}
    arctanh(B_0)=-\frac{1}{2}\ln \frac{1-B_0}{1+B_0}
\end{equation}
Thus we obtain:
\begin{equation}
    B[P_{\theta,r_e}(\cdot|q)] > tanh(\frac{\Delta |\delta r(q)|}{2} + arctanh(B_0)) - 2\delta\cdot exp(M|\delta r|)
\end{equation}
Then, note that for coefficients that are not too large, $|\delta r(q)|$ is proportional to the injected vector coefficient $r_e$ (assumption \ref{assumption:linear}), hence:
\begin{equation}
    B[P_{\theta,r_e}(\cdot|q)] > tanh(\frac{\Delta \lambda }{2}\cdot r_e + arctanh(B_0)) - 2\delta\cdot exp(M\lambda\cdot r_e)
\end{equation}
Where $\lambda$ is the coefficient relating $r_e$ to $|\delta r(q)|$. Thus for $r_e < \frac{\log \frac{\epsilon}{2\delta}}{M\cdot \lambda}$:
\begin{equation}
    B[P_{\theta,r_e}(\cdot|q)] > tanh(\frac{\Delta \lambda }{2}\cdot r_e + arctanh(B_0)) - \epsilon
\end{equation}

\section{Relation of Steering to Finetuning with Preference Learning}\label{preference}
To a degree one can draw a relation between steering and preference learning.
\begin{proposition}
    For an LLM, one iteration of gradient descent on the preference learning loss with learning rate $\eta$ is equivalent to steering with coefficient $r_e=\eta$. 
\end{proposition}

\textit{Proof:}

    The objective in preference learning is to minimize the loss:
\begin{equation}
    L = -\expectation_{(x,y^+,y^-)\sim D} [\log\frac{P(y^+|x)}{P(y^-|x)}] = -\expectation_{(x,y^+,y^-)\sim D}[\langle r^{(L)}_x, U^T(e_{y_+}-e_{y_-})\rangle]
\end{equation}
Which increases the likelihood of desired responses to prompts.
By training with preference learning, in each iteration of gradient descent, each representation is changed by:
\begin{equation}
    r^{(l)}\rightarrow r^{(l)}-\eta \frac{\partial L}{\partial r^{(l)}}
\end{equation}
The gradient of the loss \wrt~a hidden layer representation is:
\begin{equation}
    \frac{\partial L}{\partial r^l} = \expectation_{(x,y^+,y^-)\sim D}[\frac{\partial r(x)}{\partial r^l(x)} \cdot U^T(e_{y_+}-e_{y_-})]
\end{equation}
Thus at each layer, the representation is shifted in a direction that maximizes the difference between positive and negative responses' representations, $U^T(e_{y_+}-e_{y_-})$.
Which is equivalent to steering with coefficient $r_e=\eta$, and vectors $R_e=\{\expectation_{(x,y^+,y^-)\sim D}[\frac{\partial r(x)}{\partial r^l(x)} \cdot U^T(e_{y_+}-e_{y_-})]\}_{l=1}^L$

\section{Extension of Results Beyond Binary Behavior Score}\label{sec:beyond_binary}
The idea behind theorem \ref{theorem:1}, is that the resulting change to the final hidden layer due to the representation injections linearly classifies aligned and misaligned answers, where the aligned/misaligned labels are given by the binary behavior scoring function. To extend beyond a binary behavior score, we need to assume that the model's latent space captures more finegrained differences between answers. Here we will provide results for a trinary behavior score (theorem \ref{theorem:1_trinary}), and a general behavior score (theorem \ref{theorem:1_general}).

A natural extension is for a trinary score function, where $\pm 1$ is aligned/misaligned, and $0$ is irrelevant/neutral. We can reformulate theorem \ref{theorem:1} in the following way:
\begin{theorem}\label{theorem:1_trinary}
Let $P_{\theta,r_e}(\cdot|q)$ be a model prompted with query $q$ and injected with representations of coefficient $r_e$. Let  $B:\Sigma^*\rightarrow \{-1,0,+1\}$ be a behavior scoring function.
The injections to all layers amounts to a change in the final hidden layer representation that is $q$ dependent, denoted by the vector $\delta r^{(L)}_e(q)$. Assume that the representations of aligned and misaligned/irrelevant answers \wrt~$B$ are linearly separable, and that $\delta r^{(L)}_e(q)$ linearly classifies them with margin $\Delta$. Then, the behavior expectation of the model conditioned on the query $q$ satisfies:
\begin{equation}
    B[P_{\theta,r_e}(\cdot|q)] \geq  \frac{B_0 + P_+ (e^{\Delta \lambda \cdot r_e}-1)}{1 + P_+ (e^{\Delta \lambda \cdot r_e}-1)}
\end{equation}
Where $B_0 = B[P_\theta(\cdot|q)]$ and $P_+$ are the behavior expectation and probability of aligned answer without steering, and $\lambda$ is a model dependent coefficient relating between $r_e$ and the corresponding final hidden state norm.
\end{theorem}
The behavior bound has a different form, but it behaves the same -- for $r_e=0$, it coincides with $B_0$, around $r_e=0$ it is linear, and for $r_e\rightarrow\infty$ it approaches $+1$.
The proof, presented in \ref{proof:trinary}, essentially follows the proof of theorem \ref{theorem:1}, except besides the $P_\pm$ terms (probability mass of positive and negative responses without steering) there is also a $P_0$ term.

For a general behavior scoring function, $B:\Sigma^*\rightarrow [-1,+1]$, we can similarly assume that the representations of answers with score $>b_+$ and answers with score $<b_+$, are linearly separable, and obtain the following result:
\begin{theorem}\label{theorem:1_general}
Let $P_{\theta,r_e}(\cdot|q)$ be a model prompted with query $q$ and injected with representations of coefficient $r_e$. Let  $B:\Sigma^*\rightarrow [-1,+1]$ be a behavior scoring function.
The injections to all layers amounts to a change in the final hidden layer representation that is $q$ dependent, denoted by the vector $\delta r^{(L)}_e(q)$. Assume that the representations of answers with behavior score $>b+$ and those with score $<b_+$ \wrt~$B$ are linearly separable, and that $\delta r^{(L)}_e(q)$ linearly classifies them with margin $\Delta$. Then, the behavior expectation of the model conditioned on the query $q$ satisfies:
\begin{equation}
    B[P_{\theta,r_e}(\cdot|q)]\geq \frac{b_+P_+e^{\Delta\lambda r_e}-P_-}{P_+e^{\Delta\lambda r_e}+P_-}
\end{equation}
Where $P_\pm$ are the probabilities of aligned/misaligned answers without steering, and $\lambda$ is a model dependent coefficient relating between $r_e$ and the corresponding final hidden state norm.
\end{theorem}

Here we see that the behavior expectation converges to the maximal score $b_+$, for which $\delta r_e^{(L)}$ can classify answers below and above the score. The trend is similar to theorem \ref{theorem:1}, with a sigmoidal behavior, but without the tightness on behavior expectation at $r_e=0$, due to the more complex behavior scoring function. The proof is presented in \ref{proof:general}.

\subsection{Proof of theorem \ref{theorem:1_trinary}}\label{proof:trinary}
Following the same proof as in \ref{theorem:1}, up to equation \ref{eq:after_Delta}, but replacing the sum over negative answers to sum over negative and neutral answers, we obtain by denoting $P_\pm$, the sum over positive/negative answers without steering, and by $P_0$ sum over neutral answers:
\begin{equation}
    B[P_{\theta,r_e}(\cdot|q)]\geq \frac{P_+ - P_-exp(-\Delta|\delta r|)}{P_+ + (P_-+P_0)exp(-\Delta|\delta r|)}
\end{equation}
\begin{equation}
    =\frac{P_+(e^{\Delta|\delta r|}-1)+(P_+ - P_-)}{P_+(e^{\Delta|\delta r|}-1) + (P_++P_-+P_0)}
\end{equation}
We note that $P_+ + P_- + P_0=1$ and that $P_+-P_-=B[P_{\theta,r_e=0}(\cdot|q)]=B_0$:
\begin{equation}
    =\frac{P_+(e^{\Delta|\delta r|}-1)+B_0}{P_+(e^{\Delta|\delta r|}-1) + 1}
\end{equation}
Lastly, applying assumption \ref{assumption:linear}, replaces $|\delta r|=\lambda r_e$.

\subsection{Proof of theorem \ref{theorem:1_general}}\label{proof:general}
Following the same proof idea as in theorem \ref{theorem:1}, starting with equation \ref{eq:behavior_expectation} but replacing the scores in the numerator for positive and negative answers with $b_+$ and $-1$ (for worst case), up to equation \ref{eq:after_Delta}, denote by $P_+$ the probability without steering for answers with score $>b_+$ and by $P_-$ the rest:
\begin{equation}
    B[P_{\theta,r_e}(\cdot|q)]\geq \frac{b_+P_+e^{\Delta|\delta r|}-P_-}{P_+e^{\Delta|\delta r|}+P_-}
\end{equation}
Lastly, applying assumption \ref{assumption:linear}, replaces $|\delta r|=\lambda r_e$.

\clearpage
\section{Extension of Results to Multi-Token Answers}\label{sec:multi_token}
Intuitively, both the alignment guarantee result of theorem \ref{theorem:1} and helpfulness bound of theorem \ref{theorem:2}, which apply for a single token output, can be extended to multi-token answers by applying the results on multiple decoding steps.

\subsection{Alignment} Starting with alignment, we note that if the model is limited to producing $N$ tokens, then from corollary \ref{cor:1}, we can ensure that with a large enough steering coefficient, each token will correspond to an aligned response:
\begin{theorem}\label{theorem:1_multi}
        Let $\epsilon>0$, $P_\theta$ a language model, $B:\Sigma^*\rightarrow \{-1,+1\}$, behavior scoring function and $q$ a query, and suppose the model's reply contains at most $N$ tokens. Under the assumption of theorem \ref{theorem:1} holding in every decoding step, for $r_e > \frac{1}{\Delta\lambda}(\log\frac{N}{\epsilon}+\log\frac{1-B_0}{1+B_0})$, then:
    \begin{equation}
        B[P_\theta(\cdot|q)] > 1-2\epsilon
    \end{equation}
    Where $B_0$ is the behavior expectation without representation engineering.
\end{theorem}
We see that larger coefficients of steering improve the behavior expectation, similarly to corollary \ref{cor:1}, but with multiple token answers. By inverting the relation between $r_e$ and $\epsilon$, and placing it in the behavior expectation bound, we obtain a sigmoid-like behavior, that is linear for $r_e\approx0$.

\textit{Proof:}

Following the notation of the proof of theorem \ref{theorem:1}, we note that at each decoding step, the probability of outputting a token $a_i$ that is aligned \wrt~behavior scoring function $B$, conditioned on the previous context $qa_1...a_{i-1}$, is:
\begin{equation}
    \frac{\sum_{a_+\in good}P_{\theta,r_e}(a_+|qa_1...a_{i-1})}{\sum_{a_+\in good}P_{\theta,r_e}(a_+|qa_1...a_{i-1}) + \sum_{a_-\in bad}P_{\theta,r_e}(a_-|qa_1...a_{i-1})}
\end{equation}
Following the proof technique of theorem \ref{theorem:1}, we obtain that this probability is larget than:
\begin{equation}
\geq\frac{P_+e^{\Delta\lambda r_e}}{P_+e^{\Delta\lambda r_e} + P_-}
\end{equation}
Where $P_\pm$ are the probabilities for an aligned/misaligned output at the given decoding step. To ensure this probability is larger than $1-\epsilon'$, we demand:
\begin{equation}
    r_e >\frac{\log\frac{P_-}{P_+}+\log\frac{1}{\epsilon'}}{\Delta\lambda}
\end{equation}
Thus over $N$ decoding steps, we use a union bound, leading to a positive response with probability $(1-\epsilon')^N>(1-\epsilon' N)$. Taking $\epsilon' = \epsilon/N$, we obtain:
\begin{equation}
    r_e > \frac{\max_{i\in[N]}\{\log\frac{P^i_-}{P^i_+}\}+\log\frac{N}{\epsilon}}{\Delta\lambda}
\end{equation}
Where $P_\pm^i$ is the probability for a positive/negative continuation in the $i$'th token of the response. We note that $\frac{P_-^i}{P_+^i}=\frac{1-B^i_0}{1+B^i_0}$, where $B_0^i$ is the behavior expectation at the $i$'th decoding step. For the response to be positive, it is required that every step is positive, due to the binary score, then the behavior expectation of the entire response is no larger than the behavior expectation of each decoding step, $B_0 \leq \min_{i\in[N]}B_0^i$, meaning it suffices to have:  
\begin{equation}
    r_e > \frac{\log\frac{1-B_0}{1+B_0}+\log\frac{N}{\epsilon}}{\Delta\lambda}
\end{equation}

We obtain that under these conditions, an aligned response is generated with probability at least $1-\epsilon$. A negative response, is generated with probability no greater than $\epsilon$. Thus the behavior expectation is at least:
\begin{equation}
    B[P_{\theta,r_e}(\cdot|q)] > 1-2\epsilon
\end{equation}

\subsection{Helpfulness}
For helpfulness, we will consider a query $q$ and a correct answer $a$ of $N$ tokens. We will show that the probability of the answer decreases quadratically. The intuition is that in each decoding step the probability decreases quadratically, and due to the probability chain rule, if at the $i$'th step of generation, the probability for the next token is $P_i$, then the full sequence probability is $\prod_{i=1}^N P_i$. Once we expand this term \wrt~$r_e$, we get a leading quadratic dependence:
\begin{corollary}
    Let $P_\theta$ be a language model and $q$ be a query with answer $a=a_1...a_N$ containing at most $N$ tokens. Denote by $\{P_0^i\}_{i=1}^N$ the probability assigned to each correct token $\{a_i\}_{i=1}^N$ in the sequence without steering, such that the probability of the full sequence is $P_0=\prod_{i=1}^NP_0^i$. Then under the conditions of theorem \ref{theorem:2} holding at each decoding step, we have with probability of at least $1-\frac{2N}{T}$:
    \begin{equation}
        P_{\theta,r_e}(q)\leq \frac{P_0}{\prod_{i=1}^N (P_0^i +(1-P_0^i) \alpha(1-\epsilon) (1+\frac{\lambda^2 \sigma^2 \beta^2}{2}r_e^2))}
    \end{equation}
\end{corollary}
This shows the original probability of the sequence $P_0$, is normalized by a term whose leading order is quadratic in $r_e$:
\begin{equation}
    \prod_{i=1}^N (P_0^i +(1-P_0^i) \alpha(1-\epsilon) (1+\frac{\lambda^2 \sigma^2 \beta^2}{2}r_e^2))=\prod_{i=1}^N (P_0^i +(1-P_0^i) \alpha(1-\epsilon))) + c\cdot r_e^2 +o(r_e^2)
\end{equation}
We once a gain note that if $P_0^i$ is close to $1$, then $(P_0^i +(1-P_0^i) \alpha(1-\epsilon)))\approx 1$, making the bound tighter where the model is more helpful initially.

An alternative bound, is simply to consider that the probability for a sequence, $P_0$, is bounded by the probability of each element in the sequence, $P_0^i$, for which theorem \ref{theorem:2} can be directly applied, and the quadratic decay is achieved, although this is a bound that is less tight.
\end{document}